%% file: egpaper_final.tex
\documentclass[10pt,twocolumn,letterpaper]{article}

\usepackage{iccv}
\usepackage{times}
\usepackage{epsfig}
\usepackage{graphicx}
\usepackage{amsmath}
\usepackage{amssymb}
\usepackage{wrapfig}
\usepackage{graphicx}
\usepackage{amsmath}
\usepackage{amssymb}
\usepackage{booktabs}
\usepackage[dvipsnames]{xcolor}
\usepackage{xcolor,colortbl}
\usepackage{color, colortbl}
\usepackage{xcolor}
\newcommand{\tableCellHeight}{1}
\newcommand{\tabstyle}[1]{
  \setlength{\tabcolsep}{#1}
  \renewcommand{\arraystretch}{\tableCellHeight}
  \centering
  \small
}
\usepackage[accsupp]{axessibility}  
\usepackage{algorithm}
\usepackage{algorithmic}
\usepackage{sidecap}
\newcommand{\tablestyle}[2]{\setlength{\tabcolsep}{#1}\renewcommand{\arraystretch}{#2}\centering\footnotesize}
\usepackage{float}
\definecolor{purple}{RGB}{230, 227, 254}
\definecolor{lightgreen}{RGB}{238, 252, 241}
\definecolor{lightred}{RGB}{231, 187, 187}
\definecolor{darkred}{RGB}{198, 129, 129}

\definecolor{tabhighlight}{HTML}{e5e5e5}
\input{math_commands.tex}

\usepackage[font=small]{caption}
\usepackage{subcaption, multirow, overpic}
\definecolor{tabhighlight}{HTML}{e5e5e5}
\definecolor{citecolor}{HTML}{0071bc}

\def\etal{\emph{et al.}\xspace}
\def\ie{\emph{i.e.}\xspace}

\newcommand{\txt}[1]{{\texttt{#1}}}

\usepackage[algo2e]{algorithm2e}

\newcommand\extrafootertext[1]{%
    \bgroup
    \renewcommand\thefootnote{\fnsymbol{footnote}}%
    \renewcommand\thempfootnote{\fnsymbol{mpfootnote}}%
    \footnotetext[0]{#1}%
    \egroup
}

\usepackage[pagebackref=true,breaklinks=true,letterpaper=true,citecolor=blue,linkcolor=blue,colorlinks,bookmarks=false]{hyperref}

\usepackage{cleveref}
\crefformat{section}{\S#2#1#3} 
\crefformat{subsection}{\S#2#1#3}
\crefformat{subsubsection}{\S#2#1#3}

\iccvfinalcopy 

\usepackage[misc]{ifsym}
\newcommand{\corrAuthor}{$^{\textrm{\Letter}}$}

\def\iccvPaperID{12425} 

\ificcvfinal\fi

\begin{document}

\title{Self-regulating Prompts: Foundational Model Adaptation without Forgetting}

\author{%
  Muhammad Uzair Khattak$^{1*}$\corrAuthor \quad 
  Syed Talal Wasim$^{1*}$ \quad 
  Muzammal Naseer$^{1}$ \\ 
  Salman Khan$^{1,2}$ \quad
  Ming-Hsuan Yang$^{4, 5}$ \quad
  Fahad Shahbaz Khan$^{1,3}$
  \vspace{0.5em} \\
  $^{1}$Mohamed bin Zayed University of AI \quad 
  $^{2}$Australian National University \\ 
  $^{3}$Link\"{o}ping University \quad 
  $^{4}$University of California, Merced \quad 
  $^{5}$Google Research
}

\maketitle
\ificcvfinal\thispagestyle{empty}\fi

\begin{abstract}
Prompt learning has emerged as an efficient alternative for fine-tuning foundational models, such as CLIP, for various downstream tasks. Conventionally trained using the task-specific objective, i.e., cross-entropy loss, prompts tend to overfit downstream data distributions and find it challenging to capture task-agnostic general features from the frozen CLIP. This leads to the loss of the model's original generalization capability. To address this issue, our work introduces a self-regularization framework for prompting called PromptSRC (Prompting with Self-regulating Constraints). PromptSRC guides the prompts to optimize for both task-specific and task-agnostic general representations using a three-pronged approach by: (a) regulating {prompted} representations via mutual agreement maximization with the frozen model, (b) regulating with self-ensemble of prompts over the training trajectory to encode their complementary strengths, and (c) regulating with textual diversity to mitigate sample diversity imbalance with the visual branch. To the best of our knowledge, this is the first regularization framework for prompt learning that avoids overfitting by jointly attending to pre-trained model features, the training trajectory during prompting, and the textual diversity. PromptSRC explicitly steers the prompts to learn a representation space that maximizes performance on downstream tasks without compromising CLIP generalization. We perform extensive experiments on 4 benchmarks where PromptSRC  {overall} performs favorably well compared to the existing methods. Our code and pre-trained models are publicly available at: \href{https://github.com/muzairkhattak/PromptSRC}{https://github.com/muzairkhattak/PromptSRC}.
\end{abstract}\vspace{-1em}

\extrafootertext{\textsuperscript{*}Joint first authors.\\
\indent\indent \textsuperscript{\corrAuthor}\txt{uzair.khattak@mbzuai.ac.ae}}

\section{Introduction}
Vision-Language (VL) models, such as CLIP \cite{radford2021learning} and ALIGN \cite{jia2021scaling}, have demonstrated remarkable generalization capabilities for downstream tasks. These VL models are trained on large-scale web data with a contrastive loss, which allows them to encode open-vocabulary concepts by aligning pairs of images and texts in a shared embedding space. The resulting model is suited for downstream tasks such as open-vocabulary image recognition \cite{kim2022how}, object detection \cite{ feng2022promptdet}, and image segmentation \cite{luddecke2022image}.

\begin{figure*}[!ht]
\centering\vspace{-1em}
{\includegraphics[width=\textwidth]{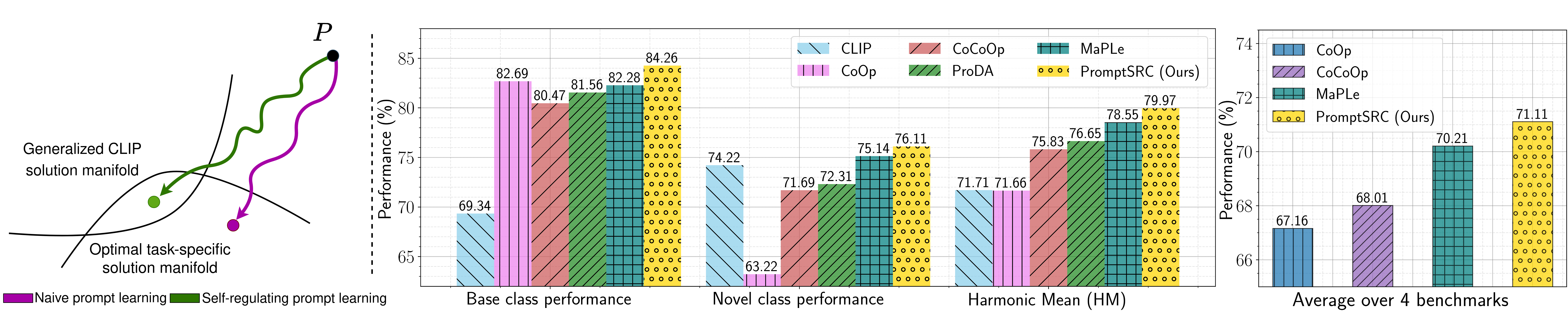}}\vspace{-0.5em}
\caption{{\color{blue}\textbf{(Left)}}: Existing prompt learning approaches rely on task-specific objectives that restrict prompt learning to learn a feature space suitable only for downstream tasks and consequently lose the generalized knowledge of CLIP (shown in \textcolor{Plum}{purple}). Our self-regulating framework explicitly guides the training trajectory of prompts towards the closest point between two optimal solution manifolds (solid line) to learn task-specific representations while also retaining generalized CLIP knowledge (shown in \textcolor{ForestGreen}{green}). {\color{blue}\textbf{(Middle)}}: Averaged across 11 image recognition datasets, PromptSRC surpasses existing methods on the base-to-novel generalization setting. {\color{blue}\textbf{(Right)}}: We evaluate our approach on four diverse image recognition benchmarks and it {overall shows competitive results} compared to the previous state-of-the-art.}  
\label{fig:concept_figure}
\end{figure*}

Prompt learning has emerged as a more efficient alternative to fine-tuning large-scale models, as shown in recent studies \cite{zhou2022conditional,zhou2022learning, chen2022prompt, huang2022unsupervised, shu2022test, lu2022prompt}. 
This approach introduces a few learnable prompt vectors to adapt models like CLIP for downstream tasks while keeping the pre-trained model weights fixed.
However, since the prompts are optimized with respect to the task-specific objective~\cite{zhou2022learning}, such as the cross-entropy loss for ImageNet~\cite{deng2009imagenet} classification, the prompted model tends to overfit to the task-specific data distribution as the training progresses. This can result in the prompted model losing the original generalization capability of the frozen CLIP model towards new tasks. Therefore, learning prompts that can model both task-specific and task-agnostic representations remain a major challenge for adapting foundational VL models.

This work seeks to self-regulate prompts to address the issue of prompt overfitting.  To this end, we propose a self-regularizing framework that guides the prompts to jointly optimize for both task-specific and task-agnostic general representations using a three-pronged approach. \textbf{a)} \emph{Regulating via Mutual Agreement Maximization:} We observe that generalizable zero-shot knowledge is preserved within frozen pre-trained VL model features but they lack task-specific knowledge. In contrast, prompts achieve better adaptation to a given task but with reduced generalizability to new tasks. Therefore, we propose to regulate learned prompts by maximizing the agreement between prompted and frozen VL model features while adapting them to the downstream task. \textbf{b)} \emph{Regulating with the Self-ensemble:} In the early epochs, prompts act are not mature to capture contextual information. As the training progresses, prompts tend to become more task-specific. Therefore we deploy a weighted prompt aggregation technique to prompts during training to regulate them using their self-ensemble over the training phase. The weights are sampled from a Gaussian distribution which suitably aggregates the useful knowledge learned by prompts at different training epochs. \textbf{c)} \emph{Regulating with Textual Diversity:} We note that unlike having multiple image samples per category for the vision encoder, there is only a single textual label available for each class. Therefore, imposing the mutual agreement constraints on multi-modal features results in sub-optimal performance due to the lack of diversity in text-side labels for the text encoder. We overcome this disparity and regulate the prompts through diverse text label templates for each class.

Overall, our approach explicitly steers prompts to learn a representation space that maximizes its performance on downstream tasks without compromising pre-trained CLIP generalization (Fig.~\ref{fig:concept_figure}: Left). We demonstrate the effectiveness of PromptSRC on four representative tasks. On the base-to-novel generalization benchmark across 11 datasets (Fig.~\ref{fig:concept_figure}: Middle), our method achieves average gains of +1.42\% in harmonic-mean over the state-of-the-art MaPLe \cite{khattak2023maple} and +8.26\% over CLIP. {Further, PromptSRC achieves competitive results in cross-dataset transfer, domain generalization, and few-shot image recognition (Fig.~\ref{fig:concept_figure}:Right)}. 

In summary, our self-regulating prompt learning framework has the following main contributions:\vspace{-0.5em}
\begin{itemize}\setlength{\itemsep}{0em}
\item We address the inherent problem of prompt overfitting for adapting foundational models through self-regularization. Our framework explicitly guides the prompts to jointly acquire both \textit{{task-specific knowledge}} and \textit{{task-agnostic generalized knowledge}} by maximizing the mutual agreement between prompted and frozen VL model features. (\cref{mutual_agreement_para})

\item  We suggest a weighted self-ensembling strategy for prompts that captures their complementary features learned at different epochs during training and enhances their generalization performance. (\cref{GPA_para})

\item To overcome the significant diversity mismatch between  the text and visual domains, we propose text-side diversity which complements limited textual labels via multiple text augmentations and regularizes prompts to learn more generalized contexts. (\cref{textual_diversity_para})
\end{itemize}

\section{Related Work}

\noindent \textbf{Vision Language models:} Foundational vision-language (VL) models \cite{radford2021learning, jia2021scaling, zhai2022lit, yao2021filip, yuan2021florence} leverage both visual and textual modalities to encode rich multi-modal representations. These models are pre-trained on a large corpus of image-text pairs available on the internet in a self-supervised manner. For instance, CLIP \cite{radford2021learning} and ALIGN \cite{jia2021scaling} utilize around 400M and 1B image-text pairs, respectively, to train their multi-modal networks. During pre-training, contrastive loss is commonly used as a self-supervision loss. This loss pulls together the features of paired images and texts while pushing away the unpaired image-text features. VL models possess a strong understanding of open-vocabulary concepts, making them suitable for various downstream vision and vision-language applications \cite{gao2021clip, zhang2021tip, rasheed2022bridging, Maaz2022Multimodal, zhou2022detecting, gu2021open, manzoor2023multimodality, zang2022open, li2022language, rao2022denseclip, ding2022decoupling}.  However, transferring these foundational models for downstream tasks without compromising on their original generalization ability still remains a major challenge. Our work aims to address this problem by proposing a novel regularization framework to adapt VL models via prompt learning.

\noindent \textbf{Prompt learning:}
Prompt learning is an alternative fine-tuning method for transferring a model towards downstream tasks without re-learning the trained model parameters. This approach adapts a pre-trained model by adding a small number of new learnable embeddings at the input known as prompt tokens. Due to its efficiency in terms of parameters and convergence rate, prompt learning is found to be of great interest for adapting  foundational models like CLIP for vision~\cite{jia2022visual,zhang2022neural,wang2022dualprompt,wang2022learning} and vision-language tasks~\cite{zhou2022learning,zhou2022conditional, zhu2022prompt, derakhshani2022variational}. CoOp \cite{zhou2022learning} fine-tunes CLIP by optimizing a continuous set of prompt vectors in its language branch for few-shot image recognition. Bahng \etal \cite{bahng2022visual} perform visual prompt tuning on CLIP by learning prompts on the vision branch. \cite{chen2022prompt} and \cite{lu2022prompt} propose to learn multiple sets of prompts for learning different contextual representations. CoCoOp \cite{zhou2022conditional} highlights the overfitting problem of CoOp and proposes to condition prompts based on visual features for improved performance on generalization tasks. {MaPLe \cite{khattak2023maple} proposes a multi-modal prompt learning approach by learning hierarchical prompts jointly at the vision and language branches of CLIP for better transfer.} Our approach builds on a variant \cite{rasheed2023fine} where prompts are learned at both the vision and language encoder of CLIP.

\noindent \textbf{Network regularization:}
Incorporating regularization techniques in neural networks has been proven to enhance their generalization capabilities \cite{lee2022cross}. Regularization strategies can be broadly classified into two streams. The first stream consists of constraint-based regularization methods, such as weight decay \cite{loshchilov2017decoupled} and adversarial training \cite{yi2021improved}. These techniques introduce additional constraints to the learning process, which helps to prevent overfitting. The second stream of regularization techniques involves modifying the inputs, model parameters, or annotations. This category includes methods such as data augmentations \cite{yun2019cutmix, zhang2017mixup, cubuk2020randaugment}, dropout \cite{srivastava2014dropout}, model ensembling \cite{ilharco2022patching, wortsman2022robust}, label smoothing \cite{szegedy2016rethinking} and batch normalization \cite{ioffe2015batch}. Our method aims to enhance the generalization performance of learned prompts via a multi-stage regularization framework, which takes inspiration from both streams of regularization techniques mentioned above. However, to the best of our knowledge, this is the first effort to regularize prompts during adaptation by jointly attending to the original VL model feature space, the training trajectory of prompts as well as the diversity of textual inputs for the multi-modal models. 

\section{Proposed Method}
\label{sec: Methodology}

\begin{figure}[!t]
    \centering
    \includegraphics[width=0.9\columnwidth]{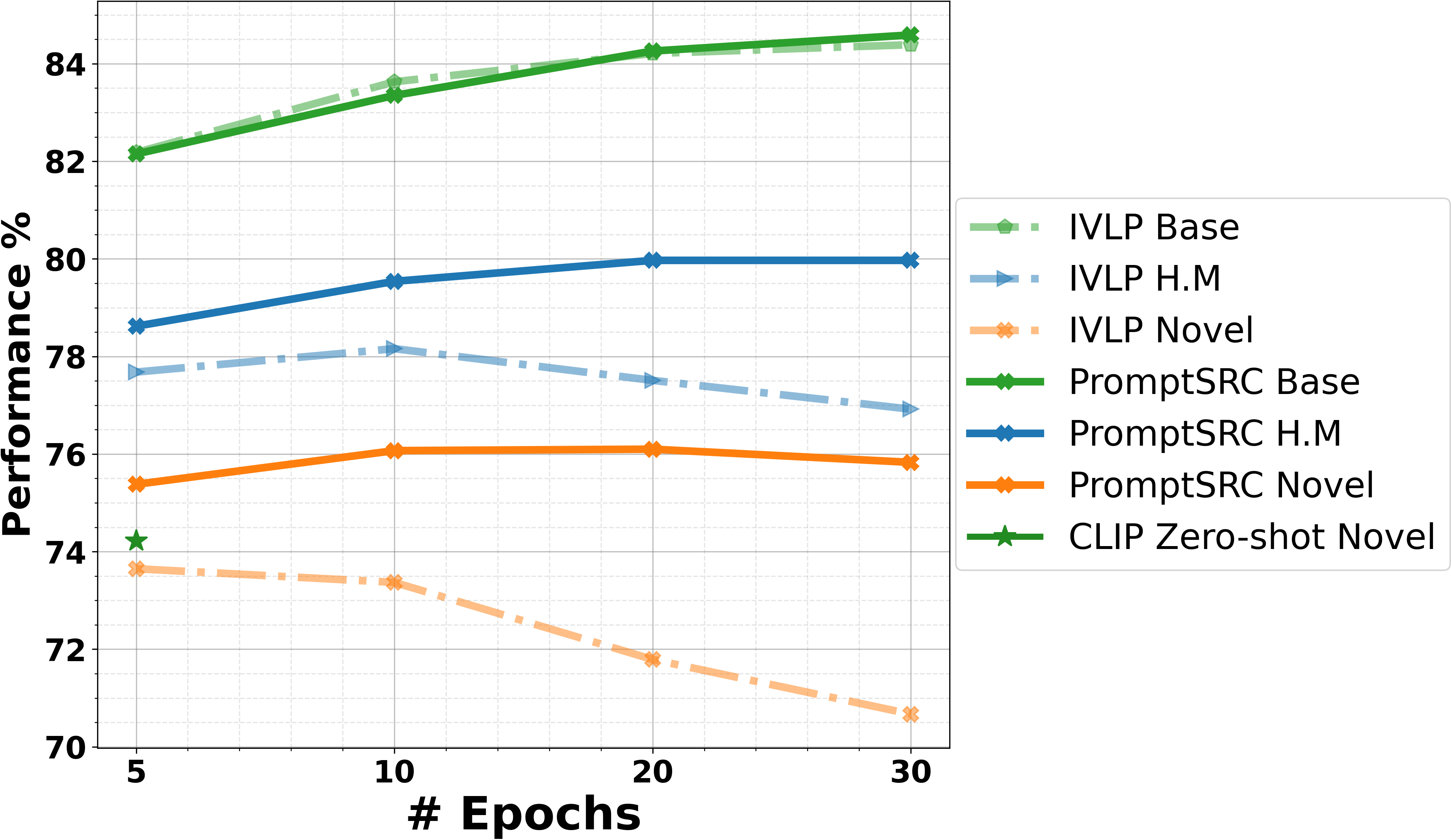}\vspace{-0.5em}
\caption{Naively training prompts with standard supervised objectives improves supervised class performance but leads to poor generalization as training schedule increases. Our PromptSRC method with explicit prompts consistency constraints improves on base classes as well as shows improvements on novel classes.}
  \label{fig:prompt_ovefitting}
\end{figure}

Prompt learning aims to adapt the general knowledge of VL foundational models like CLIP without full fine-tuning \cite{zhou2022learning, zhou2022conditional, chen2022prompt}. Since prompts are the only learnable vectors, this strategy aims to retain the pretrained generalized feature representations of CLIP while re-purposing them for downstream task-specific data via prompts. Although effective, they are susceptible to overfitting on the supervised downstream task (see Fig.~\ref{fig:prompt_ovefitting}) and their generalization towards new classes and datasets reduces as compared to the original zero-shot pre-trained CLIP. 

Our work seeks to address the overfitting behavior of prompts.
Unlike prior prompting approaches that improve generalization mainly from the model architecture perspective \cite{zhou2022conditional, khattak2023maple}, we motivate our work from the regularization perspective. 
As evidenced by the strong zero-shot performance, pre-trained CLIP features possess robust generalization characteristics. However, naively training prompts with the supervised task-specific loss struggles to retain these general attributes from the frozen CLIP. To this end, we propose a self-regularizing framework to explicitly guide the training trajectory of prompts to maximize its interaction with the pre-trained knowledge stored in the frozen CLIP.

\begin{figure*}[!ht]
\centering
{\includegraphics[width=0.85\textwidth]{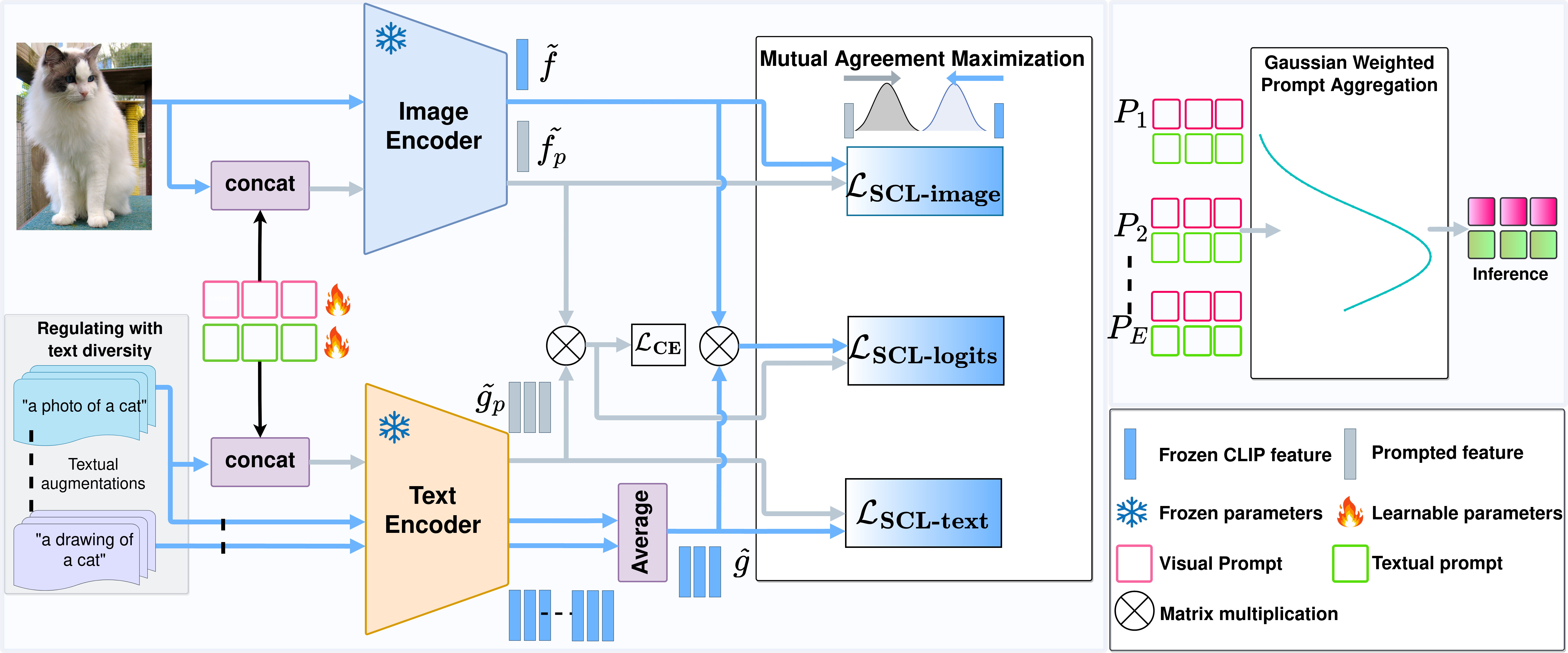}}\vspace{-0.5em}
\caption{Our proposed PromptSRC framework for self-regulating prompt learning. CLIP encoders are used to generate \textbf{\textcolor{gray}{prompted}} ($\bm{\Tilde{f}_p}$, $\bm{\Tilde{g}_p}$) and \textbf{\textcolor{RoyalBlue}{pre-trained}} ($\bm{\Tilde{f}}$, $\bm{\Tilde{g}}$) features at the image and text sides. First, we introduce textual diversity (\cref{textual_diversity_para}) and define textual augmentations to produce a diverse set of frozen VL textual features, which are averaged to represent the pre-trained VL text features ($\bm{\Tilde{g}}$). Next, we employ Mutual Agreement Maximization constraints ($\mathcal{L_{\text{SCL}}}$) to regulate the prompts, which ensure that the prompted features align well with the pre-trained VL representations at both the feature and logit levels (\cref{mutual_agreement_para}). As CLIP is frozen, we use the same VL encoders to obtain both types of features. Further, our prompt self-ensembling combines the strengths of prompts learned at different epochs ($P_1, P_2 \cdots P_E$) during training via Gaussian weighted sampling (\cref{GPA_para}). The ensembled \textbf{\textcolor{Magenta}{visual}} and \textbf{\textcolor{LimeGreen}{textual}} prompts are then used for the final inference.}
\label{fig:main_figure}
\end{figure*}

Fig. \ref{fig:main_figure} shows our overall methodology which optimizes the prompts as follows. \textbf{a)} \emph{Regularization through mutual agreement maximization:} We impose an explicit consistency constraint between prompted features and the pre-trained CLIP features within the CLIP embedding space. \textbf{b)} \emph{Regularization through prompt self-ensembling:} To further reduce overfitting, we propose a Gaussian weighted average of the prompt vectors learned at different training epochs. This ensemble-level regularization aggregates information from learned prompts across different epochs for improved generalization. \textbf{c)} \emph{Regularization through textual diversity:} Unlike having multiple images for each class, the text labels during fine-tuning are limited and bounded by the number of class categories. We incorporate textual augmentations by defining multiple text label templates for a given class. The ensemble of textual labels regularizes the prompts for better generalization during optimization.

We now continue by explaining our methodology in detail. We first revisit CLIP and CLIP-based prompt learning in \autoref{sec:meth:revisiting_clip}. This is followed by the explanation of our self-regulating prompt learning approach in \autoref{sec:meth:self_regularization_prompt_learning}.

\subsection{Preliminaries}
\label{sec:meth:revisiting_clip}

We denote the CLIP image and text encoders as ${f}$ and ${g}$, respectively and their pretrained parameters as ${\theta}_{\mathtt{CLIP}} = \{\theta_{f}, \theta_{g} \}$ where $\theta_{f}$ and $\theta_{g}$ refer to the image and text encoder parameters, respectively. The input image $\bm{X} \in \mathbb{R}^{C\times H\times W}$ is divided into $M$ patches followed by a projection to produce patch tokens.  Further, a learnable class token $\bm{e}_{cls}$ is appended with the input patches as $\bm{\Tilde{X}}= \{\bm{e}_{cls}, \bm{e}_{1}, \bm{e}_{2}, \cdots, \bm{e}_{M}\}$.
The image encoder ${f}$ encodes the input patches via multiple transformer blocks to produce a latent visual feature representation $\bm{\Tilde{f}} = f(\bm{\Tilde{X}}, \theta_{f})$, where $\bm{\Tilde{f}} \in  \mathbb{R}^{d}$. Next, the corresponding class label ${y}$ is wrapped within a text template 
such as ‘a photo of a \{class label\}’ which can be formulated as $\bm{\Tilde{Y}}=\{\bm{t}_{SOS}, \bm{t}_{1}, \bm{t}_{2}, \cdots, \bm{t}_{L}, \bm{c}_{k}, \bm{t}_{EOS}\}$. Here $\{\bm{t}_l|_{l=1}^{L}\}$ and $\bm{c}_{k}$ are the word embeddings corresponding to the text template 
and the class label, respectively while $\bm{t}_{SOS}$  and  $\bm{t}_{EOS}$ are the learnable start and end token embeddings. The text encoder ${g}$ encodes $\bm{\Tilde{Y}}$ via multiple transformer blocks to produce the latent textual feature as $\bm{\Tilde{g}} = g(\bm{\Tilde{Y}}, \theta_{g})$, where $\bm{\Tilde{g}} \in  \mathbb{R}^{d}$. For zero-shot inference, textual features of text template 
with class labels $\{1, 2, \cdots, C\}$
are matched with image feature $\bm{\Tilde{f}}$ as $\frac{\mathtt{exp}(\mathtt{sim}(\bm{\Tilde{g}} \cdot \bm{\Tilde{f}})\tau)}{\sum_{i=1}^{C}\mathtt{exp}(\mathtt{sim}(\bm{\Tilde{g_i}} \cdot \bm{\Tilde{f}})\tau)}$, where $\mathtt{sim}()$ denotes the cosine similarity and $\tau$ is the temperature.

\noindent \textbf{Prompt Learning for CLIP:} Prompt learning approaches append learnable prompt tokens at either the text \cite{zhou2022learning, zhou2022conditional} encoder or image \cite{bahng2022visual} encoder. We use a simple baseline method \cite{rasheed2023fine} that learns hierarchical prompt tokens on both the text and image encoders separately, named as Independent Vision-Language Prompting (IVLP).

Specifically, we append learnable $T$ language and $V$ visual prompts given as $\bm{P_{t}} = \{\bm{p_t}^1,\bm{p_t}^2, \cdots, \bm{p_t}^T\}$ and $\bm{P_{v}} = \{\bm{p_v}^1,\bm{p_v}^2, \cdots, \bm{p_v}^V\}$ with the textual and visual input tokens, respectively. Therefore, the image encoder processes the following input tokens $\bm{\Tilde{X}_p}= \{ \bm{P_{v}}, \bm{e}_{cls}, \bm{e}_{1}, \bm{e}_{2}, \cdots, \bm{e}_{M}\}$ to generate prompted visual feature represented as $\bm{\Tilde{f}_p} = f(\bm{\Tilde{X}_p}, \theta_{f})$. Similarly, textual feature is obtained as $\bm{\Tilde{g}_p} = g(\bm{\Tilde{Y}_p}, \theta_{g})$, where $\bm{\Tilde{Y}_p}=\{\bm{t}_{SOS}, \bm{P_{t}} ,\bm{t}_{1}, \bm{t}_{2}, \cdots, \bm{t}_{L}, c_{k}, \bm{t}_{EOS}\}$. In contrast to shallow prompting where learnable prompts are introduced only at the first transformer block of the image and text encoders, our approach uses deep prompting which learns separate sets of prompts at every transformer block.   The vision and language  prompts are jointly represented as $ \bm{P} = \{ \bm{P_{v}}, \bm{P_{t}}\}$. The feature representations obtained using these learnable prompts are referred to as \emph{prompted features}. 

For image classification on a downstream dataset $\mathcal{D}$, prompts $\bm{P}$ interact with pre-trained and frozen $\theta_{f}$ and $\theta_{g}$ and are optimized with the cross-entropy loss, $\mathcal{L_{\text{CE}}}$, as:
\begin{align}
\label{eq:LCE}
    \mathcal{L_{\text{CE}}} = \text{arg}&\min_{\bm{P}}\mathbb{E}_{(\bm{X}, {y})\sim\mathcal{D}} \, \mathcal{L} (\text{sim}(\bm{\Tilde{f}_p},\bm{\Tilde{g}_p}), y).
\end{align}

\subsection{Self-Regularization for Prompt Learning}
\label{sec:meth:self_regularization_prompt_learning}

The $\mathcal{L_{\text{CE}}}$ objective employs ground truth labels to optimize the prompts for the downstream task.
As a result, the prompts adapt and learn \emph{task-specific knowledge}. During training, prompts interact with pre-trained and frozen CLIP tokens through self-attention layers in the transformer blocks. This interaction of prompts tokens with pre-trained CLIP weights ${\theta}_{\mathtt{CLIP}}$ provides implicit regularization and encourages retaining the \emph{task-agnostic generalized knowledge} within learned prompts.
However, as shown in Fig.~\ref{fig:prompt_ovefitting}, prompts tend to overfit on the supervised task and drift away from the generalized CLIP space as the training schedule increases. Consequently, new task performance is degraded, despite the fact that CLIP image and text encoder weights $\theta_{f}$ and $\theta_{g}$ are kept frozen. 
As prompts undergo further training, the implicit generalization constraint becomes weaker against the task-specific $\mathcal{L_{\text{CE}}}$ objective.

One naive approach to address this issue is to reduce the training schedule to balance the performance between the base and new tasks. However, training the prompts for fewer iterations to prevent losing generalization comes at the cost of relatively lower performance on the supervised task. Here, we present a prompt learning approach that maximizes supervised task performance without sacrificing performance on novel tasks and classes. We propose to anchor prompt training with self-regularization which constitutes three main components as discussed below.

\subsubsection{Mutual agreement maximization}
\label{mutual_agreement_para}
As discussed above, the strong downstream dataset transfer constraint imposed by $\mathcal{L_{\text{CE}}}$ causes the prompts to overfit on task-specific data and it struggles to effectively utilize the general information from the frozen CLIP. We propose to explicitly guide the training trajectory by imposing a constraint to maximize its mutual agreement between the prompted and the frozen CLIP features. We achieve this by explicitly conditioning the prompted features to be consistent with the CLIP features obtained without learnable prompts. As we do not require any second model for such conditioning, we call this regularizing constraint as a self-consistency loss (SCL).
For a given input sample and its corresponding textual label, we obtain visual features using learnable prompts and pre-trained visual features, $\bm{\Tilde{f}_{p}}$ and $\bm{\Tilde{f}} $ within the frozen CLIP latent space. Similarly, we obtain textual features $\bm{\Tilde{g}_{p}}$ and $\bm{\Tilde{g}}$.

We then impose a constraint on the prompted visual and text features to ensure their consistency with the CLIP pre-trained features as follows,
\begin{align}
\label{eq:scl-features}
    \mathcal{L_{\text{SCL-image}}} = \sum_{i=1}^{d}|\bm{\Tilde{f}_{p}} - \bm{\Tilde{f}}|, \; 
    \mathcal{L_{\text{SCL-text}}} = \sum_{i=1}^{d} |\bm{\Tilde{g}_{p}} - \bm{\Tilde{g}}|.
\end{align}
As shown in Eq. \ref{eq:scl-features}, we utilize $L1$ loss to impose the feature level consistency. Note that our self-consistency constraint is also compatible with other variants of matching losses such as cosine similarity or MSE loss which we study in our ablations (\autoref{sec:experiments:ablation_experiments}).

To further complement the regularization constraint and maximize the alignment between the general features and the prompted features, we impose logit level self-consistency regularization and condition the prompted logits distribution on pre-trained CLIP logits distribution by minimizing the Kullback-Leibler divergence as follows,
\begin{align}
\label{eq:scl-logits}
    \mathcal{L_{\text{SCL-logits}}} = \mathcal{D_{KL}} (\text{sim}(\bm{\Tilde{f}_p},\bm{\Tilde{g}_{p}}), {\text{sim}(\bm{\Tilde{f}},\bm{\Tilde{g}}))}.
\end{align}

Overall, the self-consistency training objectives guide the prompts to gain complementary  knowledge from pre-trained CLIP features, therefore providing strongly generalized prompts,
\begin{align}
\label{eq:scl-combined}
    \mathcal{L_{\text{SCL}}} &= \lambda_{1}\mathcal{L_{\text{SCL-image}}} + \lambda_{2}\mathcal{L_{\text{SCL-text}}} + \mathcal{L_{\text{SCL-logits}}},
\end{align}
where $\lambda_{1}$ and $\lambda_{2}$ are loss balancing hyper-parameters. Our overall training objective thus becomes,
\begin{align}
\label{eq:final-loss}
    \mathcal{L_{\text{final}}} &= \mathcal{L_{\text{CE}}} + \mathcal{L_{\text{SCL}}} .
\end{align}

\noindent \textbf{Discussion on $\mathcal{L_{\text{final}}}$:}
$\mathcal{L_{\text{SCL}}}$ loss guides the prompts to converge at solutions that are generalized. On the other hand, $\mathcal{L_{\text{CE}}}$ guides the prompts to maximize performance on the downstream supervised tasks. The combination of these losses conditions the prompts to maximize their performance on supervised tasks and at the same time guides the prompts learning trajectory toward a weight space that is consistent with the CLIP zero-shot features. As shown in Fig. \ref{fig:prompt_ovefitting}, our proposed methodology maximizes the supervised tasks' performance while also improving the generalization. This shows that the proposed training objectives for prompt learning setup are complementary to each other. 

\subsubsection{Regularization with prompt self-ensembling}
\label{GPA_para}
The second component in our self-regularizing framework enforces regularization using prompt self-ensembling. Model ensembling in the weight space has been shown to improve both the performance and generalization of a model \cite{wortsman2022robust, ilharco2022patching}. However, it has not been actively studied in the context of prompt learning, where prompts are only learnable parameters with frozen model parameters. 

To effectively utilize the prompts knowledge from the previous training iterations, we propose prompts aggregation for a generalizable solution. For a training schedule with $E$ total epochs, prompts at every epoch are given by $\{{\bm{P}\}}_{t=1}^{E}$. Aggregated prompts (AP) are then calculated as,
\begin{equation}
\label{eq:p-tuning}
    \{\bm{P}\}^{\text{AP}} = \sum_{t=1}^{E}{\frac{w_{t} . \bm{P}}{\sum_{i=1}^{E}w_{i}}},
\end{equation}
where $w_{i}$ is the weight assigned to prompts at each epoch $t$. 

In the early epochs, prompts are not mature to capture contextual information due to their random initialization. During aggregation, they should be given less weight as they act as noise which is carried along with the input tokens. On the other hand, the prompts learned in the last few epochs are task specific and highly favours the supervised downstream task distribution. We propose to perform Gaussian weighted prompt aggregation (GPA), where small aggregation weights are given to prompts at initial epochs, higher weights to prompts at middle epochs, and relatively lower weights to prompts at final epochs, resulting in optimal prompt representations that improve generalization to downstream tasks. GPA provides optimal weight values $w_{i}$ by sampling from a 
Gaussian distribution $w_{i} \sim \mathcal{N}(\mu,\,\sigma^{2})$, where $\sigma^{2}$ and $\mu$ are hyper-parameters and $\sum_{i=1}^{E}w_{i}=1$. Gaussian distribution is defined over the epochs and its mean is dictated by the epoch number. We formulate this weighting as a moving average to avoid saving multiple copies of prompts by keeping one additional copy which is updated via aggregation at every epoch $i$,
\begin{equation}
\label{eq:moving_average}
    {\bm{P}}^{\text{GPA}} =  \sum_{i=1}^{E}w_{i} . \bm{P_{i}}.
\end{equation}

\subsubsection{Regulating prompts with textual diversity}
\label{textual_diversity_para}
Through the $\mathcal{L_{\text{SCL}}}$ loss, the visual prompted features to instill \textit{diverse generalized contexts} from pre-trained CLIP visual features as multiple image samples are present for each label category. This provides a natural source of augmentations at the image side and promotes additional regularization. However, as opposed to having multiple images per category, we note that the text space during fine-tuning is limited, and prompted features are learned based on pre-trained CLIP text features, with only one feature representation per category. This mismatch between the available diversity at the image and text side leads to sub-optimal learning of prompted textual features. To address the diversity mismatch, we incorporate textual diversity in the text encoder. Specifically, we use a pool of textual prompt templates $\{PT|_{l=1}^{N}\}$, containing $N$ augmentations to form multiple text features per category. The pre-trained CLIP textual features are now obtained as an ensemble of multiple prompts templates $\bm{\Tilde{g}} = \frac{1}{N} \sum_{i=1}^{N} \bm{\Tilde{g}^{i}}$. As pre-trained CLIP textual features are now represented by the ensemble of multiple augmentations for each label, the prompted textual features learn more \textit{diverse generalized contexts} from the frozen CLIP. We note that the proposed textual diversity is different from the standard prompt ensembling technique explored by CLIP authors. CLIP uses ensemble of text prompts during inference for classification. In contrast, we utilize them during training for self-regularization by enforcing mutual agreement of ensembled features with prompted features, and prompted features are used at inference. Next, we show the efficacy of our proposed components via comprehensive experiments provided below.

\input{tables/main_experiments/base2novel_new_table}

\section{Experiments}
\subsection{Evaluation settings}
We extensively evaluate our approach and present a comparison with other methods on four benchmark settings. \\
\textbf{Base-to-novel class generalization:} In this setting, we equally split the datasets into base and novel classes. The model is trained on base classes and evaluated on both base classes and novel classes. This benchmark evaluates the generalization ability of a method within a dataset. 

\noindent \textbf{Few-shot learning:} We incorporate this setting to compare the learning capacity of the model under extremely limited supervision and verify if our approach learns {complementary} task-specific and task-agnostic knowledge. For each dataset, we test the model's generalization for different $K$-shots per category, where $K = 1, 2, 4, 8, 16$.  

\noindent \textbf{Domain generalization setting:} We train a source model on ImageNet~\cite{deng2009imagenet} and evaluate on out-of-distribution datasets to test performance under domain shifts. 

\noindent \textbf{Cross-dataset evaluation:} In cross-dataset transfer, we train the models on ImageNet \cite{deng2009imagenet} and directly evaluate it on other datasets without any data-specific fine-tuning. 

\noindent \textbf{Datasets:} For base to novel class generalization, few-shot setting and cross-dataset evaluation, we follow CoOp \cite{zhou2022learning} and CoCoOp \cite{zhou2022conditional}, and use 11 image recognition datasets. The datasets cover multiple recognition tasks including ImageNet~\cite{deng2009imagenet} and Caltech101~\cite{fei2004learning} which consists of generic objects; OxfordPets~\cite{parkhi2012cats}, StanfordCars~\cite{krause20133d}, Flowers102~\cite{nilsback2008automated}, Food101~\cite{bossard2014food}, and FGVCAircraft~\cite{maji2013fine} for fine-grained classification, SUN397~\cite{xiao2010sun} for scene recognition, UCF101~\cite{soomro2012ucf101} for action recognition, DTD~\cite{cimpoi2014describing} for texture classification, and EuroSAT~\cite{helber2019eurosat} which consists of satellite images. For domain generalization benchmark, we use ImageNet~\cite{deng2009imagenet} as a source dataset and use ImageNet-A~\cite{hendrycks2021natural}, ImageNet-R~\cite{hendrycks2021many}, ImageNet-Sketch~\cite{wang2019learning} and ImageNetV2~\cite{recht2019imagenet} as out of distribution datasets.

\noindent \textbf{Implementation details:} We use a ViT-B/16 based CLIP model in our experiments and report results averaged over 3 runs. We use deep prompting with $V=T=4$ VL prompts and train for 50 epochs for few-shot setting and 20 epochs the rest of the 3 benchmarks respectively.  For domain generalization and cross-dataset evaluation, we train the ImageNet source model on all classes with $K=16$ shots using $V=T=4$ VL prompts in the first 3 transformer layers. For few-shot and base-to-novel setting, prompts are learned in the first 9 transformer layers. Prompts are randomly initialized with a normal distribution except the text prompts of the first layer which are initialized with the word embeddings of ``a photo of a". We fix the learning rate to 0.0025. We set $\lambda_{1}=10$ and $\lambda_{2}=25$ to weight $\mathcal{L_{\text{SCL-image}}}$ and $\mathcal{L_{\text{SCL-text}}}$ respectively. The corresponding hyperparameters are fixed across all datasets and benchmarks. For textual diversity, we use a total of $N=60$ standard prompt templates provided in \cite{radford2021learning}. For comparison with ProDA \cite{lu2022prompt}, we report their results produced by \cite{derakhshani2022variational}. Refer to Appendix \ref{appendix:additional_implementation_details} for additional implementation details.
\subsection{Effectiveness of Self-regulating Prompts}
We first disentangle the regularization components in our self-regulating prompting framework and show the individual contributions in Table \ref{tab:ablations_on_components}.
Baseline IVLP provides high base class performance but suffers from poor generalization (row-1). By enforcing mutual agreement through $\mathcal{L}_\text{SCL}$ (row-2), novel class performance significantly increases by 3.95\% while maintaining base class gains. This suggests that $\mathcal{L}_\text{SCL}$ explicitly enforces the prompts to capture the generalizable features from frozen CLIP. 
Integrating GPA (row-3) which suitably aggregates prompts across the training cycle further reduces overfitting and improves the novel class performance. Finally, combined with textual diversity to overcome the diversity mismatch between the text and visual domains (row-4), PromptSRC achieves improvements on both base and novel classes, leading to the average novel class and harmonic mean gains of +4.31\% and +2.46\% respectively. The averaged results on 11 datasets are summarized in Table \ref{tab:ablations_on_components}. Note that even small improvements in these metrics correspond to significant gains. We refer the readers to Appendix \ref{appendix:base2novel_per_component_per_dataset} for results on individual datasets.

\begin{figure*}[!ht]
\centering
{\includegraphics[width=1\textwidth]{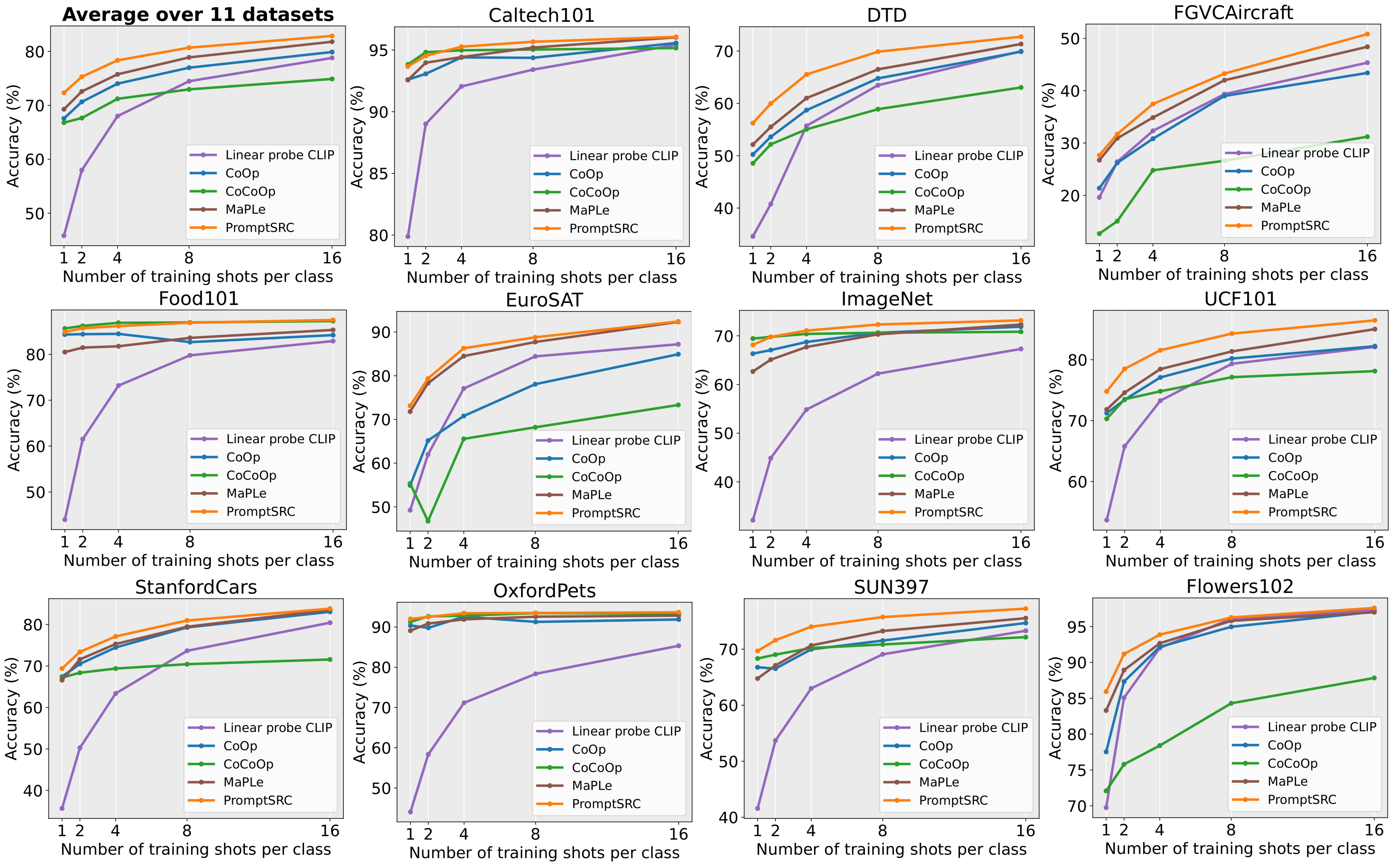}}\vspace{-0.5em}
   \caption{\textnormal{PromptSRC performance comparison in few-shot image recognition setting.} All methods are trained on ViT-B/16 CLIP backbone using their best settings. PromptSRC demonstrates consistent improvements over existing methods specifically for lesser shots \ie $K=1,2,4$. On average, PromptSRC provides the highest performance gains for all shots. These results demonstrate that PromptSRC learns complementary task-agnostic general features from frozen CLIP without being restricted from learning downstream task representations.}
\label{fig:few_shot_comparision}
\end{figure*}

\input{tables/ablations/component_wise_ablation}
\subsection{Base-to-Novel Generalization}
We compare the performance of our approach with zero-shot CLIP \cite{radford2021learning}, CoOp \cite{zhou2022learning}, CoCoOp \cite{zhou2022conditional}, ProDA \cite{lu2022prompt} and MaPLe \cite{khattak2023maple}, in Table \ref{tab:base-to-new}. Overall, all existing approaches outperform zero-shot CLIP on base classes but show inferior performance on novel classes {except MaPLe}. This suggests that they {overall} tend to lose the generalizable features stored in the frozen CLIP model. In contrast, PromptSRC significantly improves base class performance while improving the zero-shot CLIP novel class accuracy by 1.88\%. This shows the importance of explicit guidance provided by PromptSRC in learning complementary task-specific and task-agnostic representations which aid base and novel classes respectively.

CoOp is heavily trained on base classes and consequently compromises on its generalization. For instance, on EuroSAT \cite{helber2019eurosat}, CoOp provides a substantial 92.19\% base class accuracy and inferior novel class accuracy of 54.74\%. On the other hand, PromptSRC which learns self-regulating prompts provides the highest base and novel class accuracies of 92.90\% and 73.90\% on EuroSAT respectively.

In comparison to CoCoOp and ProDA, PromptSRC shows gains on the 10/11 datasets respectively. {Against the recent MaPLe approach, PromptSRC improves performance on 8/11 datasets while using 77x less tunable parameters (3.55M of MaPLe vs 46K of PromptSRC).} With respect to the averaged results, PromptSRC provides the best results of 84.26\%, 76.10\%, and 79.97\%  on the base class, novel class, and harmonic mean respectively. 

\subsection{Few-shot Experiments}
To explicitly verify if our regularization framework restricts the prompts to learn task-specific knowledge or not, we compare our few-shot results with existing methods in Fig. \ref{fig:few_shot_comparision}.
In general, all prompt learning approaches perform better than the linear probe, especially in scenarios with lesser shots \ie, $K=1, 2, 4$. PromptSRC overall provides consistent improvements on all shots in comparison with all existing methods. {When compared with the existing best method MaPLe, PromptSRC consistently provides absolute gains of 3.05\%, 2.72\%, 2.59\%,	1.80\%, and, 1.07\% on 1, 2, 4, 8, and 16 shots respectively which are averaged over 11 datasets.} 
Furthermore, we note that our approach achieves relatively larger gains in minimal data cases such as for $K=1, 2$ for almost all datasets. This demonstrates that PromptSRC regulates prompts against overfitting without restricting the prompts to learn task-specific knowledge. 

\input{tables/main_experiments/cross_dataset.tex}

\subsection{Cross Dataset Evaluation}
We compare our cross-dataset performance with previous methods in Table \ref{tab:xd}. On the source dataset, PromptSRC performs comparably to other methods. In comparison with CoOp and CoCoOp, PromptSRC shows competitive performance and achieves better generalization in 8/10 and 7/10 datasets respectively. {Compared with MaPLe, PromptSRC shows improved performance in 5/10 datasets while utilizing significantly less tunable parameters (46K vs 3.55M).} 

\subsection{Domain Generalization Experiments}
Table \ref{tab:robustness} summarizes the results of PromptSRC and previous methods on out-of-distribution datasets. We directly evaluate our model trained on ImageNet. On target datasets, PromptSRC consistently outperforms all existing methods, with an overall highest average accuracy of 60.65\%. This suggests that our self-regulating framework favors better generalization for datasets with domain shifts.
\input{tables/main_experiments/dg.tex}

\subsection{Ablative Analysis}
\label{sec:experiments:ablation_experiments}

\noindent \textbf{Embedding consistency loss ablation:}
In Table \ref{table:matching_loss_ablations}, we ablate on the choice of matching loss metric used in our proposed feature level $\mathcal{L_{\text{SCL}}}$ loss constraints. For simplicity, we only incorporate $\mathcal{L_{\text{SCL-image}}}$ and $\mathcal{L_{\text{SCL-text}}}$ on top of the IVLP baseline. Generally, distance-based matching metrics outperform the cosine similarity metric in terms of generalization as they impose a much harder constraint. Overall, the $L1$ matching metric provides the highest HM.

\noindent \textbf{Prompt ensembling:}
Table \ref{tab:ablations_on_ensembling} shows ablation on various prompt ensembling techniques. Using equal weights for prompts reduces base class results as initial epoch prompts are not mature enough. In contrast, our proposed Gaussian weighted prompt aggregation results in the highest performance. Detailed ablation experiments for other hyper-parameters are provided in Appendix \ref{appendix:additional_ablation}.
\input{tables/ablations/scl_loss_ablations}
\input{tables/ablations/prompt_ensemble_ablation}

\input{tables/supplementory/compute_cost_analysis}

\noindent \textbf{{Training and inference compute cost analysis:}} In Table \ref{tab_appendix:compute_analysis}, we show the compute cost analysis of our approach and compare it with other prompting methods. PromptSRC's overall training GFLOPs are only 0.13x higher than baseline IVLP, while it maintains the same GFLOPs and throughput during inference. Pre-trained CLIP textual features are pre-computed and a single additional forward pass is required through image encoder to compute pre-trained CLIP visual features for our mutual agreement maximization technique. Training time of PromptSRC is 9.3\% longer than IVLP which is significantly lower than CoCoOp. We use 4 vision and text prompts similar to the IVLP.

\begin{figure}[!t]
    \centering
    \includegraphics[width=\columnwidth]{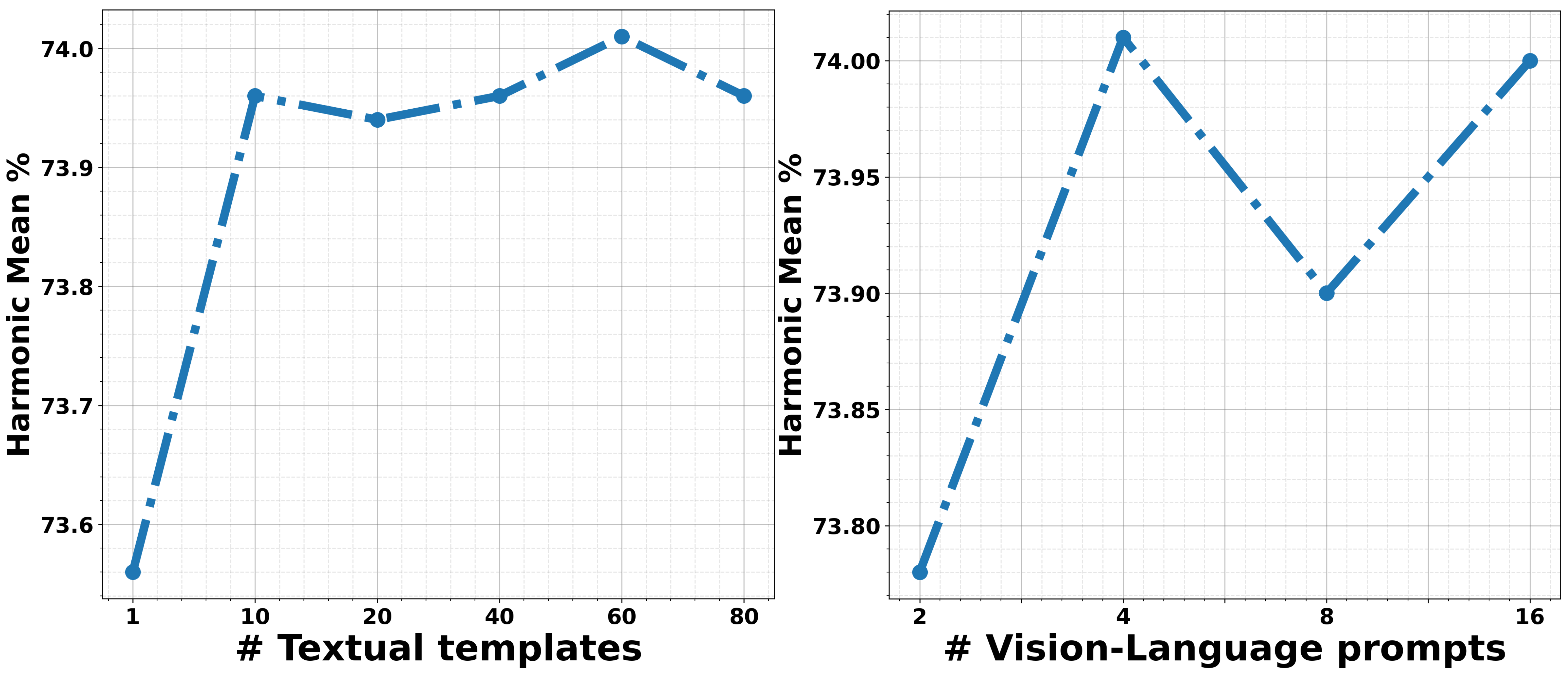}
\caption{Ablation study on the number of textual prompts for textual diversity (left) and prompt token length (right) on ImageNet.}
  \label{fig:ctx_templates_ablation}
\end{figure}

\noindent \textbf{Prompt Length:}
Fig. \ref{fig:ctx_templates_ablation} (right) shows the effect of prompt token length on the harmonic mean. Overall, the performance increases as prompt length increases. Using 4 vision-language prompts provides the highest harmonic mean.

\noindent \textbf{No. of templates in textual diversity:}
In Fig. \ref{fig:ctx_templates_ablation} (left), we ablate on the number of text prompt templates for textual diversity. We note that increasing the number of textual templates for textual diversity generally increases the performance. This suggests that adding textual diversity using multiple templates for pre-trained features provides more rich supervision for the learned prompted features.

\section{Conclusion}\vspace{-0.5em}
Prompt learning has emerged as an effective paradigm for adapting foundational VL models like CLIP. However, the prompts learned by the majority of existing methods inherently tend to overfit task-specific objectives and consequently compromise the inherent generalization ability of CLIP. Our work proposes a self-regulating prompt learning framework that addresses the prompt overfitting problem for better generalization. We show it is critical to guide the training trajectory of prompts by explicitly encouraging its mutual agreement with the frozen model through self-consistency constraints supplemented by incorporating textual diversity. We also propose a self-ensembling strategy for prompts that appropriately aggregates them via a Gaussian-weighted approach over the course of training. Extensive evaluations on multiple benchmarks show the benefit of our self-regulating approach for prompt learning.

{\small
\bibliographystyle{ieee_fullname}
\bibliography{egbib}
}

\newpage

\appendix

\begin{center}
\textbf{{\Large Supplementary Material}\\[0.5em]
{\large Self-regulating Prompts: Foundational Model Adaptation without Forgetting}
}
\end{center}\vspace{1em}

The following section contains supplemental information and encompasses more implementation details, results comparison, and a thorough ablative analysis of PromptSRC. The contents are organized in the following order.
\begin{itemize}
    \item Additional implementation details (Appendix~\ref{appendix:additional_implementation_details})
    \item Additional results comparison (Appendix~\ref{appendix:base2novel_per_component_per_dataset})
    \item Additional ablative analysis (Appendix~\ref{appendix:additional_ablation})
\end{itemize}

\section{Additional Implementation details}
\label{appendix:additional_implementation_details}
\noindent \textbf{{Additional Training details:}} We use a publically available ViT-B/16 CLIP model with $d=512$ and use a learning rate of 0.0025 which is fixed for all experiments in all benchmarks. We train PromptSRC for 50 epochs for few-shot settings and 20 epochs for the remaining three benchmark settings respectively. The respective epochs are fixed across all datasets. All models are trained using SGD optimizer and utilize a single NVIDIA A100 GPU. 

\noindent \textbf{Gaussian Weighted Prompt Aggregation (GPA):}
We note that the prompts learned in the initial training epochs are not mature and act as noise due to their random initialization. 
On the other hand, prompts learned in the last few epochs are task-specific and highly favors the supervised downstream task distribution. GPA strives to maintain a balance by assigning lower weights to initial prompts, higher weights to middle prompts, and relatively lower weights to final prompts, resulting in optimal prompt representations that improve generalization to downstream tasks. Gaussian distribution in GPA is defined over the epochs and its mean is dictated by the epoch number. We then sample weights ($w_{i} \sim \mathcal{N}(\mu,\,\sigma^{2})$) for prompts of every epoch to get the final prompt aggregation. Hyper-parameters are set using validation splits
Table \ref{tab_appendix:GPA_hyper_parameters} shows the hyper-parameter values chosen for the proposed GPA technique, which are kept fixed for respective base-to-novel generalization, cross-dataset and domain generalization setting. For few-shot setting, we use $\mu=30$ and $\sigma^2=30$ for ImageNet, Caltech101, OxfordPets, Food101, UCF101 and  SUN397. For datasets including StanfordCars, Flowers102, FGVCAircraft, DTD and EuroSAT, we use $\mu=45$ and $\sigma^2=5$.

\input{tables/supplementory/GPA_hyper_paramters}

\noindent \textbf{Textual diversity:}
For the textual diversity technique, we randomly select 60 prompt templates from the complete template list provided in \cite{radford2021learning}. Specifically, our textual diversity component uses the following prompt templates. \\  {
    \texttt{``a photo of a $\{$category$\}$."}\\
    \texttt{``a bad photo of a $\{$category$\}$."}\\
    \texttt{``a photo of many $\{$category$\}$."}\\
    \texttt{``a sculpture of a $\{$category$\}$."}\\
    \texttt{``a photo of the hard to see $\{$category$\}$."}\\
    \texttt{``a low resolution photo of the $\{$category$\}$."}\\
    \texttt{``a rendering of a $\{$category$\}$."}\\
    \texttt{``graffiti of a $\{$category$\}$."}\\
    \texttt{``a bad photo of the $\{$category$\}$."}\\
    \texttt{``a cropped photo of the $\{$category$\}$."}\\
    \texttt{``a tattoo of a $\{$category$\}$."}\\
    \texttt{``the embroidered $\{$category$\}$."}\\
    \texttt{``a photo of a hard to see $\{$category$\}$."}\\
    \texttt{``a bright photo of a $\{$category$\}$."}\\
    \texttt{``a photo of a clean $\{$category$\}$."}\\
    \texttt{``a photo of a dirty $\{$category$\}$."}\\
    \texttt{``a dark photo of the $\{$category$\}$."}\\
    \texttt{``a drawing of a $\{$category$\}$."}\\
    \texttt{``a photo of my $\{$category$\}$."}\\
    \texttt{``the plastic $\{$category$\}$."}\\
    \texttt{``a photo of the cool $\{$category$\}$."}\\
    \texttt{``a close-up photo of a $\{$category$\}$."}\\
    \texttt{``a black and white photo of the $\{$category$\}$."}\\
    \texttt{``a painting of the $\{$category$\}$."}\\
    \texttt{``a painting of a $\{$category$\}$."}\\
    \texttt{``a pixelated photo of the $\{$category$\}$."}\\
    \texttt{``a sculpture of the $\{$category$\}$."}\\
    \texttt{``a bright photo of the $\{$category$\}$."}\\
    \texttt{``a cropped photo of a $\{$category$\}$."}\\
    \texttt{``a plastic $\{$category$\}$."}\\
    \texttt{``a photo of the dirty $\{$category$\}$."}\\
    \texttt{``a jpeg corrupted photo of a $\{$category$\}$."}\\
    \texttt{``a blurry photo of the $\{$category$\}$."}\\
    \texttt{``a photo of the $\{$category$\}$."}\\
    \texttt{``a good photo of the $\{$category$\}$."}\\
    \texttt{``a rendering of the $\{$category$\}$."}\\
    \texttt{``a $\{$category$\}$ in a video game."}\\
    \texttt{``a photo of one $\{$category$\}$."}\\
    \texttt{``a doodle of a $\{$category$\}$."}\\
    \texttt{``a close-up photo of the $\{$category$\}$."}\\
    \texttt{``the origami $\{$category$\}$."}\\
    \texttt{``the $\{$category$\}$ in a video game."}\\
    \texttt{``a sketch of a $\{$category$\}$."}\\
    \texttt{``a doodle of the $\{$category$\}$."}\\
    \texttt{``a origami $\{$category$\}$."}\\
    \texttt{``a low resolution photo of a $\{$category$\}$."}\\
    \texttt{``the toy $\{$category$\}$."}\\
    \texttt{``a rendition of the $\{$category$\}$."}\\
    \texttt{``a photo of the clean $\{$category$\}$."}\\
    \texttt{``a photo of a large $\{$category$\}$."}\\
    \texttt{``a rendition of a $\{$category$\}$."}\\
    \texttt{``a photo of a nice $\{$category$\}$."}\\
    \texttt{``a photo of a weird $\{$category$\}$."}\\
    \texttt{``a blurry photo of a $\{$category$\}$."}\\
    \texttt{``a cartoon $\{$category$\}$."}\\
    \texttt{``art of a $\{$category$\}$."}\\
    \texttt{``a sketch of the $\{$category$\}$."}\\
    \texttt{``a embroidered $\{$category$\}$."}\\
    \texttt{``a pixelated photo of a $\{$category$\}$."}\\
    \texttt{``itap of the $\{$category$\}$."}\\
}

\noindent  \textbf{{Evaluation metrics:}}
We report top-1 base-class and novel-class accuracy for each dataset in base-to-novel generalization setting. We also report harmonic mean (HM) between base and novel class accuracy which is the main metric that represents generalization performance.

For all shots ($K=1, 2, 4, 8, 16$) in few-shot setting, we report top-1 accuracies obtained on the corresponding test-set of each dataset using the splits provided in CoOp \cite{zhou2022learning}. 

Similar to few-shot setting, we report top-1 accuracies obtained on the test set of each dataset for cross dataset evaluation and domain generalization experiments respectively.

\noindent  \textbf{{Algorithm:}}
In \autoref{alg:example}, we show the pseudo-code implementation of our proposed PromptSRC framework.

\begin{algorithm}[!t]\small
   \caption{Learning Self-regulating prompts}
   \label{alg:example}
\begin{algorithmic}
   \STATE {\bfseries Input:} Dataset $\mathcal{D}=\{X, y\}^{N}$, Model ${\theta}_{\mathtt{CLIP}} = \{\theta_{g} , \theta_{f} \}$, Prompt vectors $ \bm{P} = \{ \bm{P_{v}}, \bm{P_{t}}\}$. No. of text templates = $N$. iteration (i) = 1.
   \STATE {\bfseries Require:} Initialize GPA prompt param. $ \bm{P}^{\text{GPA}} = \{ \bm{p_{v}}, \bm{p_{t}}\}^{GPA}$. Sample Gaussian weights for GPA $\{w_{1}, w_{2}, w_{3}, \cdot, w_{T} \}$. GPA is applied after every $c$ iterations.
   \FOR{\(\text{i}\in[1,T]\)}
   \STATE sample data $\{X, y\} \subseteq \mathcal{D}$ \\
   \tcp{prompted features.}
   \STATE Using ${\theta}_{\mathtt{CLIP}}$ and $\bm{P}$, obtain prompted visual and text features $\bm{\Tilde{f_p}} \leftarrow  f(\bm{\Tilde{x}_p}, \theta_{{f}}),~ \bm{\Tilde{g}_p} \leftarrow  g(\bm{\Tilde{y}_p}, \theta_{g})$  \\
   \tcp{normal CE supervision loss.}
   \STATE $\mathcal{L_{\text{sup}}} \leftarrow \mathcal{L_{\text{CE}}}(\text{sim}(\bm{\Tilde{f}_p},\bm{\Tilde{g}_{p}}), y)$ \\
   \tcp{pre-trained features.}
   \STATE Obtain pre-trained visual and textual features using only ${\theta}_{\mathtt{CLIP}}$  $\bm{\Tilde{f}} \leftarrow  f(\bm{\Tilde{x}}, \theta_{{f}}),~ \bm{\Tilde{g}} \leftarrow \frac{1}{N} \sum_{i=1}^{N} g(\bm{\Tilde{y}^{i}}, \theta_{g})$  \\ 
      \tcp{self-regularizing consistency losses.}
      \STATE $\mathcal{L_{\text{SCL}}} \leftarrow \lambda_{1}\mathcal{L_{\text{SCL-image}}}(\bm{\Tilde{f}_{p}},\bm{\Tilde{f}}) + \lambda_{2}\mathcal{L_{\text{SCL-text}}}(\bm{\Tilde{g}_{p}},\bm{\Tilde{g}}) + \mathcal{L_{\text{SCL-logits}}}(\text{sim}(\bm{\Tilde{f}_p},\bm{\Tilde{g}_{p}}), \text{sim}(\bm{\Tilde{f}},\bm{\Tilde{g}}))$ \\
    \tcp{compute total loss.}
    \STATE $\mathcal{L_{\text{final}}} \leftarrow  \mathcal{L_{\text{sup}}} +  \mathcal{L_{\text{SCL}}} $ \\
   \tcp{update prompt vectors with combined loss.}
    $ \bm{P} \leftarrow \bm{P} -  \delta \nabla_{\bm{P}} \mathcal{L_{\text{final}}} $
   \quad \\
   \tcp{Gaussian prompt ensembling.}
   \IF{$\text{mod}(\text{i}, c) == 0$}
   \STATE $\bm{P}^{\text{GPA}} \leftarrow \bm{P}^{\text{GPA}} + w_{\text{i}} . \bm{P}$
   \ENDIF
   \ENDFOR
\end{algorithmic}
\end{algorithm}

\section{Additional results comparison}
\label{appendix:base2novel_per_component_per_dataset}
In this section, we provide additional per-dataset results comparison and show the compatibility of PromptSRC for diverse tasks and recent VL models.

\noindent \textbf{{Generalization of PromptSRC towards video understanding tasks:}} We verify the applicability of our approach across new tasks and evaluate PromptSRC on a video action recognition generalization benchmark. Following the base-to-novel generalization setting of ViFi-CLIP \cite{rasheed2023fine}, we employ PromptSRC on a Kinetics-400 pre-trained ViFi-CLIP \cite{rasheed2023fine} and learn prompts on UCF-101 video dataset. The results are shown in Table \ref{tab_appendix:video_task}. In comparison with the naive IVLP method, PromptSRC shows favorable performance gains and even surpasses fully fine-tuned video-adapted CLIP models like ActionCLIP. This suggests that the proposed PromptSRC approach can generalize to other diverse modality downstream tasks including videos.

\input{tables/supplementory/video_task}
\input{tables/supplementory/alternate_textual_diversity}

\input{tables/supplementory/eva_clip_comparision}

\noindent \textbf{{Compatibility of PromptSRC in recent foundational VL models:}} We have demonstrated the effectiveness of our approach on the CLIP  Vision-Language (VL) model in the main manuscript. To assess how our approach scales with more recent foundational VL models, we conduct analysis using a newly introduced VL model, EVA-CLIP (CVPR'23) \cite{fang2023eva}. EVA-CLIP has been pre-trained using advanced self-supervision and optimization techniques. We employ the IVLP and PromptSRC prompting approaches to fine-tune the EVA-CLIP ViT-B/16 model in the base-to-novel generalization setting. The comparison of results is presented in Table \ref{tab_appendix:eva_clip_comparison}. PromptSRC consistently improves the generalization performance on 10/11 datasets and provides an absolute average HM gain of +2.09\% in comparison with the IVLP baseline approach.

\noindent \textbf{{Results of individual components:}} In Table \ref{tab_appendix:base_to_new}, we show the per-dataset results for each component of our PromptSRC framework in the base-to-novel generalization setting. Our results indicate that overall, the proposed regularization components are effective in improving performance in comparison with the naive IVLP prompt learning approach.

\section{Additional ablation study}
\label{appendix:additional_ablation}
\noindent \textbf{On Variants of Textual diversity:} Our proposed method for achieving textual diversity involves using an ensemble of frozen CLIP textual features obtained through multiple text augmentations. Here, we provide an analysis of an alternate approach for incorporating textual diversity. Instead of using an ensemble, we use a single prompt template chosen at random from N available templates to generate frozen CLIP textual features. The results averaged over 11 datasets, are shown in Table \ref{tab_appendix:alternate_textual_diversity}. However, we observe that PromptSRC with the ensembled textual diversity technique outperforms the alternate approach. This suggests that using an ensemble of frozen CLIP features encourages the learning of more diverse prompt representations.

Below, we conduct detailed ablation experiments on the ImageNet validation set to analyze the effect of GPA hyper-parameters on the final performance.

\begin{figure}[!t]
    \centering
    \includegraphics[width=\columnwidth]{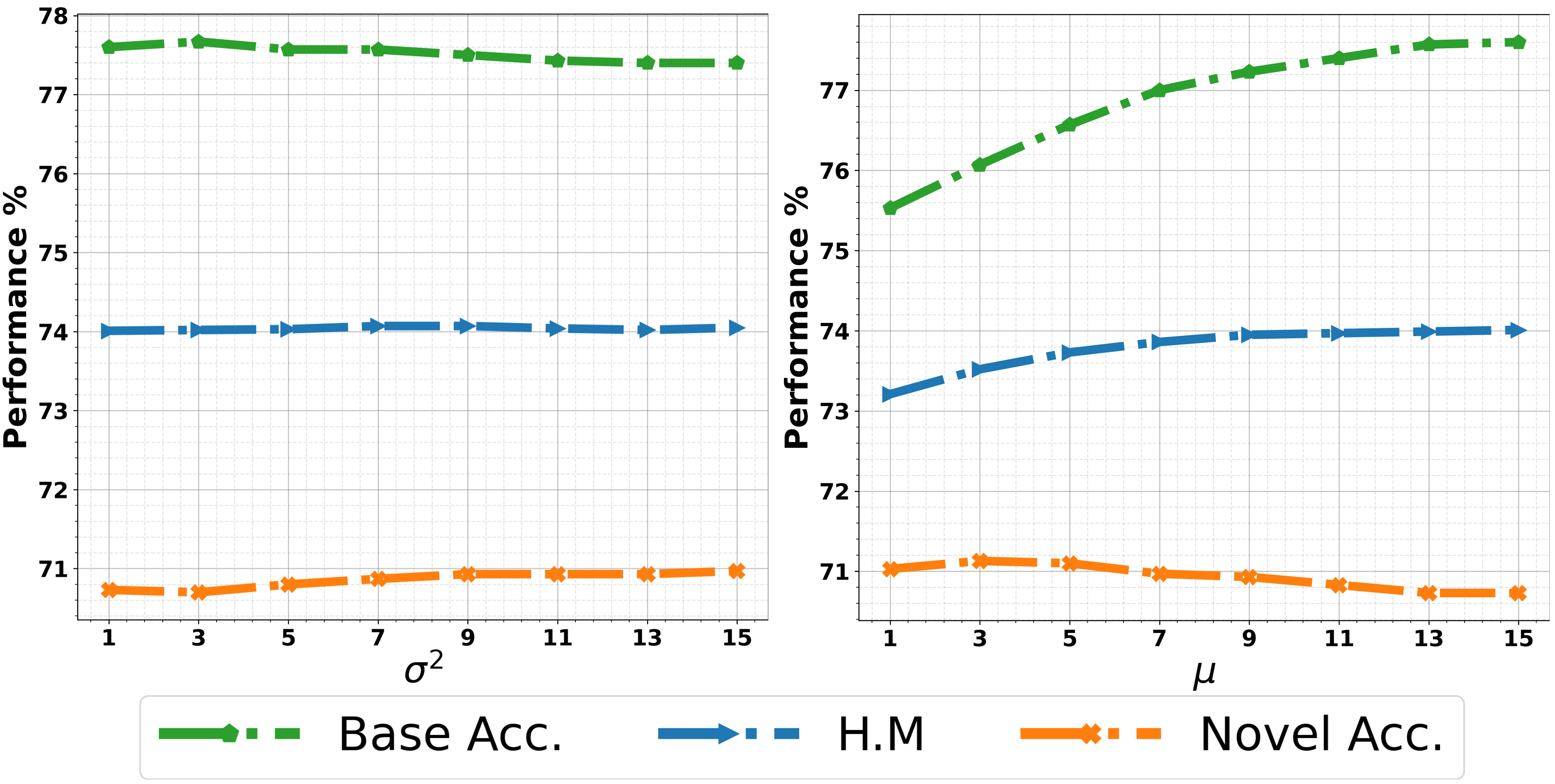}
\caption{Ablation on GPA hyper-parameters on ImageNet.}
  \label{fig:gpa_hyperparameters}
\end{figure}

\noindent \textbf{GPA hyper-parameters:}
We conduct ablation on $\mu$ and $\sigma^{2}$ hyper-parameters of GPA for the ImageNet dataset and show the results in Figure \ref{fig:gpa_hyperparameters}. Overall, varying $\sigma^{2}$ has a minute effect on performance. On the other hand, as we increase $\mu$, GPA provides more weights to prompts learned in the latter epochs which increases the base class performance and slightly decreases the novel class performance.

\input{tables/supplementory/per_dataset_result_individual_components}

\input{tables/supplementory/few_shot_per_dataset}
 \noindent \textbf{{Few-shot experiments:}} Table \ref{tab_appendix:few_shot_experiments} shows the detailed per-dataset results of various methods in the few-shot setting. Overall, PromptSRC achieves consistant improvements over existing methods for all shots.

\end{document}

%% file: math_commands.tex
\usepackage{amsmath,amsfonts,bm}

\def\eqref#1{equation~\ref{#1}}

\def\1{\bm{1}}

\DeclareMathAlphabet{\mathsfit}{\encodingdefault}{\sfdefault}{m}{sl}
\SetMathAlphabet{\mathsfit}{bold}{\encodingdefault}{\sfdefault}{bx}{n}



%% file: tables/main_experiments/base2novel_new_table.tex
\begin{table}[t!]
\centering
\tablestyle{-12pt}{1.1}
\addtolength{\tabcolsep}{+14pt}
\resizebox{\columnwidth}{!}{%
\begin{tabular}{lc|c c c c c c c}
\toprule
{Dataset} &  &  {CLIP} &  {CoOp} & {CoCoOp} &  {ProDA} & 
{MaPLe} & 
{PromptSRC} &  {$\Delta$}  \\
 & &  \cite{radford2021learning} & \cite{zhou2022learning} & \cite{zhou2022conditional}  & \cite{lu2022prompt}  & \cite{khattak2023maple}  & (Ours) & \\
\midrule
\multirow{3}{*}{\shortstack[l]{Average on\\  11 datasets}}       & Base      & 69.34  & 82.69   & 80.47  & 81.56   & 82.28        & \textbf{84.26}  &  \textcolor{MidnightBlue}{{+2.0}}\\
                               & Novel     & 74.22  & 63.22   & 71.69  & 72.30    & 75.14          & \textbf{76.10}  &   \textcolor{MidnightBlue}{{+1.0}}\\
                               & HM        & 71.70  & 71.66   & 75.83 &76.65     & 78.55        & \textbf{79.97}  &   \textcolor{MidnightBlue}{{+1.4}}\\
\midrule
\multirow{3}{*}{ImageNet}      & Base   & 72.43  & 76.47   &  75.98 &  75.40     & 76.66         & \textbf{77.60}  &  \textcolor{MidnightBlue}{{+0.9}}\\ 
                               & Novel    & 68.14  & 67.88   &  70.43 & 70.23      & 70.54       & \textbf{70.73}  &   \textcolor{MidnightBlue}{{+0.2}}\\ 
                               & HM          & 70.22  & 71.92   &  73.10 & 72.72   & 73.47     &\textbf{74.01}  &  \textcolor{MidnightBlue}{{+0.5}}\\
\midrule
\multirow{3}{*}{Caltech101}    & Base   & 96.84  & 98.00   &  97.96 & \textbf{98.27}   & 97.74          & 98.10 &   \textcolor{MidnightBlue}{{+0.4}}\\
                               & Novel     & 94.00  & 89.81   &  93.81 & 93.23   & \textbf{94.36}  & {94.03} & \textcolor{Bittersweet}{{-0.3}}\\
                               & HM          & 95.40  & 93.73   &  95.84 &  95.68  & \textbf{96.02} & \textbf{96.02} &  \textcolor{MidnightBlue}{{+0.0}}\\
\midrule
\multirow{3}{*}{OxfordPets}    & Base     & 91.17  & 93.67   & 95.20 & \textbf{95.43}       & \textbf{95.43 }      & {95.33}  &  \textcolor{Bittersweet}{{-0.1}}\\
                               & Novel      & 97.26  & 95.29   & {97.69} & \textbf{97.83}      & 97.76       & 97.30  &   \textcolor{Bittersweet}{{-0.5}}\\
                               & HM          & 94.12  & 94.47   & {96.43} & \textbf{96.62}      & 96.58        & 96.30  &   \textcolor{Bittersweet}{{-0.3}}\\
\midrule
\multirow{3}{*}{\shortstack[l]{Stanford \\ Cars}}  & Base    & 63.37 & 78.12    & 70.49 &74.70      & 72.94         & \textbf{78.27}  &  \textcolor{MidnightBlue}{{+5.3}}\\
                               & Novel     & 74.89 & 60.40    & 73.59& 71.20        & 74.00      & \textbf{74.97}  &  \textcolor{MidnightBlue}{{+1.0}}\\
                               & HM         & 68.65 & 68.13    & 72.01  &  72.91        & 73.47         & \textbf{76.58}  &  \textcolor{MidnightBlue}{{+3.1}}\\
\midrule
\multirow{3}{*}{Flowers102}    & Base     & 72.08 & 97.60    & 94.87  &97.70       & 95.92      &\textbf{98.07}  &  \textcolor{MidnightBlue}{{+2.1}}\\
                               & Novel     & \textbf{77.80} & 59.67    & 71.75 &68.68     & 72.46        & 76.50  & \textcolor{MidnightBlue}{{+4.1}}\\
                               & HM         & 74.83 & 74.06    & 81.71  & 80.66    & 82.56      & \textbf{85.95}  & \textcolor{MidnightBlue}{{+3.4}}\\
\midrule
\multirow{3}{*}{Food101}       & Base     & 90.10 & 88.33    & {90.70}  & 90.30    & \textbf{90.71 }      & 90.67  &\textcolor{Bittersweet}{{-0.1}}\\
                               & Novel     & 91.22 & 82.26    & 91.29  & 88.57       & \textbf{92.05}     &{91.53}  & \textcolor{Bittersweet}{{-0.5}}\\
                               & HM          & 90.66 & 85.19    & 90.99  & 89.43    & \textbf{91.38}   & {91.10}  & \textcolor{Bittersweet}{{-0.3}}\\
\midrule
\multirow{3}{*}{\shortstack[l]{FGVC \\ Aircraft}}  & Base    & 27.19 & 40.44    & 33.41 & 36.90    & 37.44       & \textbf{42.73}  &  \textcolor{MidnightBlue}{{+5.3}}\\
                               & Novel      & 36.29 & 22.30    & 23.71 &34.13      & 35.61      & \textbf{37.87}  &  \textcolor{MidnightBlue}{{+2.3}}\\
                               & HM         & 31.09 & 28.75    & 27.74  & 35.46    & 36.50   & \textbf{40.15}  &  \textcolor{MidnightBlue}{{+3.7}}\\
\midrule
\multirow{3}{*}{SUN397}        & Base    & 69.36 & 80.60    &  79.74  &78.67       & 80.82    &\textbf{82.67 } &   \textcolor{MidnightBlue}{{+1.9}}\\
                               & Novel      & 75.35 & 65.89    &  76.86 &76.93    & \textbf{78.70 }        & {78.47}  &  \textcolor{Bittersweet}{{-0.2}}\\
                               & HM        & 72.23 & 72.51    &  78.27 &77.79   & 79.75    & \textbf{80.52}  &  \textcolor{MidnightBlue}{{+0.8}}\\
\midrule
\multirow{3}{*}{DTD}           & Base    & 53.24 & 79.44    & 77.01 &80.67        & 80.36      & \textbf{83.37}  &  \textcolor{MidnightBlue}{{+3.0}}\\
                               & Novel     & 59.90 & 41.18    & 56.00 &56.48      & 59.18       & \textbf{62.97 } &  \textcolor{MidnightBlue}{{+3.8}}\\
                               & HM          & 56.37 & 54.24    & 64.85 &66.44    & 68.16      & \textbf{71.75}  & \textcolor{MidnightBlue}{{+3.6}}\\
\midrule
\multirow{3}{*}{EuroSAT}       & Base    & 56.48 & 92.19    & 87.49  &83.90        & \textbf{94.07}       & {92.90}  &  \textcolor{Bittersweet}{{-1.2}}\\
                               & Novel      & 64.05 & 54.74    & 60.04& 66.00      & 73.23      & \textbf{73.90}  & \textcolor{MidnightBlue}{{+0.7}}\\
                               & HM          & 60.03 & 68.69    & 71.21& 73.88  & \textbf{82.35}   & {82.32}  &  \textcolor{Bittersweet}{{-0.1}}\\
\midrule
\multirow{3}{*}{UCF101}        & Base     & 70.53  & 84.69   & 82.33  &85.23         & 83.00       &\textbf{ 87.10}  & \textcolor{MidnightBlue}{{+4.1}}\\
                               & Novel      & 77.50  & 56.05   & 73.45  &71.97     &78.66         & \textbf{78.80}  &  \textcolor{MidnightBlue}{{+0.1}}\\
                               & HM          & 73.85  & 67.46   & 77.64 &78.04   & 80.77    & \textbf{82.74}  & \textcolor{MidnightBlue}{{+2.0}}\\
\bottomrule
\end{tabular}%
}\vspace{-0.5em}
    \caption{\small\textnormal{Accuracy comparison on Base-to-novel generalization of PromptSRC with previous methods}. The prompts learned with our self-regularizing approach show overall consistent improvements on base classes, without losing generalization. Absolute gains over MaPLe \cite{khattak2023maple} are shown
    in \textcolor{MidnightBlue}{blue}.}
    \label{tab:base-to-new}
\end{table}

%% file: tables/ablations/component_wise_ablation.tex
\begin{table}[!t]

    \small \centering
 \setlength{\tabcolsep}{8pt}
    \scalebox{0.85}[0.85]{
    \begin{tabular}{l cc | c }
    \toprule
    Method  & Base Acc. & Novel Acc. & HM\\
    \midrule
    1: Independent V-L prompting & 84.21 & 71.79 & 77.51 \\
    2: + $\mathcal{L}_\text{SCL}$ & 84.21 & 75.38 & 79.55 \\
    3: + GPA & 84.16	& 75.69	& 79.70 \\
    \rowcolor{tabhighlight}
    4: + Textual diversity & \textbf{84.26} & \textbf{76.10} & \textbf{79.97} \\
    \bottomrule
    \end{tabular}
    }\vspace{-0.5em}
    \caption{Effect of our proposed regularization techniques. Results are averaged over 11 datasets. HM refers to harmonic mean. 
    }
    \label{tab:ablations_on_components}
\end{table}

%% file: tables/main_experiments/cross_dataset.tex
\begin{table}[!t]
    \tabstyle{2.5pt}
    \scalebox{0.65}{
    \begin{tabular}{l c ccccccccccc}
    \toprule
    & \textbf{Source} & \multicolumn{11}{c}{\textbf{Target}} \\ \cmidrule(lr){2-2} \cmidrule(lr){3-13}
    & \rotatebox{90}{ImageNet} & \rotatebox{90}{Caltech101} & \rotatebox{90}{OxfordPets} & \rotatebox{90}{StanfordCars} & \rotatebox{90}{Flowers102} & \rotatebox{90}{Food101} & \rotatebox{90}{Aircraft} & \rotatebox{90}{SUN397} & \rotatebox{90}{DTD} & \rotatebox{90}{EuroSAT} & \rotatebox{90}{UCF101} & \rotatebox{90}{\emph{Average}} \\
    \midrule
    CoOp & \textbf{71.51} & 93.70 & 89.14 & 64.51 & 68.71 & 85.30 & 18.47 & 64.15 & 41.92 & {{46.39}} & 66.55 & 63.88 \\
    Co-CoOp & 71.02 &\textbf{ 94.43} & {90.14} & 65.32 & {71.88} & 86.06 & 22.94 & \textbf{67.36} & 45.73 & 45.37 & 68.21 & 65.74 \\
    MaPLe & 70.72 & 93.53 & \textbf{{90.49}} & {65.57} &\textbf{ {72.23}} & \textbf{{86.20}} & \textbf{{24.74}} & 67.01 & {46.49} & {\textbf{{48.06}}} & {68.69} & \textbf{{66.30}}  \\
    \midrule
\rowcolor{tabhighlight} PromptSRC & 71.27 & \textbf{93.60} & {{90.25}} & \textbf{65.70} & {70.25} & {{86.15}} & {{23.90}} & 67.10 & \textbf{46.87} & {{45.50}} & \textbf{68.75} & {{65.81}} \\
    \bottomrule
    \end{tabular}}\vspace{-0.5em}
        \caption{\textnormal{Cross-dataset benchmark evaluation.} PromptSRC achieves overall favourable performance.}
    \label{tab:xd} 
\end{table}


%% file: tables/main_experiments/dg.tex
\begin{table}[!t]

    \small \centering
 \setlength{\tabcolsep}{8pt}
    \scalebox{0.75}[0.75]{
    \begin{tabular}{l cccccc}
    \toprule
    & \textbf{Source} & \multicolumn{5}{c}{\textbf{Target}} \\ \cmidrule(lr){2-2} \cmidrule(lr){3-7}
     & ImageNet & -V2 & -S & -A & -R  & Avg.\\
    \midrule
    CLIP &  66.73 & 60.83 & {46.15} & 47.77 & {73.96} & {57.18} \\
    CoOp &  \textbf{71.51} & {64.20} & 47.99  & 49.71  & 75.21  & {59.28} \\
    Co-CoOp & 71.02 & {64.07} & 48.75 & 50.63 & 76.18 & {59.91}  \\
        MaPLe & 70.72  & {64.07} & 49.15  & 50.90 & 76.98 & {60.27}  \\
    \midrule
    \rowcolor{tabhighlight} PromptSRC & 71.27 & \textbf{{64.35}} & \textbf{49.55} & \textbf{50.90}  & \textbf{77.80} & \textbf{60.65} \\
    \bottomrule
    \end{tabular}}\vspace{-0.5em}
        \caption{\textnormal{Domain generalization. }Prompt learning methods are trained on imageNet and evaluated on datasets with domain shifts.} 
    \label{tab:robustness}
    \vspace{-1em}
\end{table}

%% file: tables/ablations/scl_loss_ablations.tex
\begin{table}[t!]
    \small \centering
 \setlength{\tabcolsep}{7pt}
    \scalebox{0.75}[0.75]{
    \begin{tabular}{lccc}
    \toprule
    Method  & Base Acc. & Novel Acc. & HM \\
    \midrule
    1: Independent V-L prompting (IVLP) & 84.21 & 71.79 & 77.51 \\
    2: IVLP + Cosine similarity & 84.47	&74.51	&79.17 \\
    3: IVLP + Mean square error (MSE)  & \textbf{84.59}&	74.68 &	79.33 \\
     \rowcolor{tabhighlight} 4: IVLP + $L1$  & 84.42 & \textbf{74.99} & \textbf{79.43} \\
    \bottomrule
    \end{tabular}}
    \caption{Effect of matching losses for $\mathcal{L}_\text{SCL-image}$ and $\mathcal{L}_\text{SCL-image}$ consistency objectives. $L1$ matching loss provides highest HM.}
    \label{table:matching_loss_ablations}
\end{table}

%% file: tables/ablations/prompt_ensemble_ablation.tex
\begin{table}[t!]
   \small \centering
  \setlength{\tabcolsep}{10pt}
    \scalebox{0.75}[0.75]{
    \begin{tabular}{l cccc}
    \toprule
    Method  & Base Acc. & Novel Acc. & HM\\
            \midrule
    1: Exponential moving average & 83.09	&76.15 &	79.47 \\
    2: Equal weighting (averaging) & 83.50	&\textbf{76.47}	&79.83 \\
    \rowcolor{tabhighlight} 3:  GPA (Ours) & \textbf{84.26} & {76.10} & \textbf{79.97} \\
    \bottomrule
    \end{tabular}
    }\vspace{-0.5em}
    \caption{Ablation on prompt ensembling techniques. Gaussian weighted prompt aggregation (GPA) provides better performance. 
    }
    \label{tab:ablations_on_ensembling}
\end{table}

%% file: tables/supplementory/compute_cost_analysis.tex
\begin{table}[!t]
\tablestyle{5.5pt}{1.1}
\addtolength{\tabcolsep}{-4.4pt}
\scalebox{1}{
\begin{tabular}{lccccc}
\toprule
    Method & {GFLOP (train)} & {GFLOP (test)} & {Train time (min)} & FPS & HM\\ 
        \midrule
        CoOp   & 162.5 & 162.5& 10.08 & 1344 & 71.66\\
        CoCoOp & 162.5 &162.5&	39.53 & 15.08 & 75.83\\
        
        IVLP  & 162.8&	162.8&	12.01 & 1380 & 77.51\\
     \rowcolor{tabhighlight}   PromptSRC & 179.6	&162.8 &	13.13 & 1380 & \textbf{79.97}\\
\bottomrule
    \end{tabular}
} 
\vspace{-1em}
\caption[caption]{{\small PromptSRC compute cost comparison using SUN397 dataset. Training time for all methods is calculated for 10 epochs on a single A100 GPU on SUN397 dataset.}
\label{tab_appendix:compute_analysis}
}
\end{table}

%% file: tables/supplementory/GPA_hyper_paramters.tex
\begin{table}[!h]

    \small \centering
 \setlength{\tabcolsep}{8pt}
    \scalebox{0.75}[0.75]{
    \begin{tabular}{l ccc}
    \toprule
    GPA parameter & Base-to-Novel  & Cross dataset & D.G\\
    \midrule
$\mu$ & 15& 6 & 6 \\
$\sigma^{2}$ & 1 & 10  & 10\\
    \bottomrule
    \end{tabular}
    }\vspace{-0.5em}
    \caption{Hyper-parameters settings used in GPA technique for various benchmark settings. D.G refers to domain generalization.
    }
    \label{tab_appendix:GPA_hyper_parameters}
\end{table}

%% file: tables/supplementory/video_task.tex
\begin{table}[t!]
    \small \centering
 \setlength{\tabcolsep}{20pt}
    \scalebox{0.75}[0.75]{
    \begin{tabular}{lccc}
    \toprule
    Method  & Base Acc. & Novel Acc. & HM\\
    \midrule
    Vanilla CLIP  & 78.50 & 63.60  & 70.30 \\
    ActionCLIP &  {85.60 } & {75.30} & {{80.10}}\\
    XCLIP   & 95.40  & 74.00 & 83.40 \\
    A5 &  {95.80  } & {71.00} & {81.60}\\
    \midrule
    IVLP & 95.90 & 74.10 & 83.60 \\
   \rowcolor{tabhighlight}  PromptSRC &  \textbf{96.43 } & \textbf{76.79} & \textbf{85.50}\\
    \bottomrule
    \end{tabular}}\vspace{-0.5em}
    \caption{Performance comparison in video action recognition generalization benchmark on UCF-101. We employ PromptSRC and IVLP on ViFi-CLIP and compare with the prior video approaches.}
    \label{tab_appendix:video_task}
\end{table}

%% file: tables/supplementory/alternate_textual_diversity.tex
\begin{table}[!t]
    \small \centering
 \setlength{\tabcolsep}{6pt}
    \scalebox{0.75}[0.75]{
    \begin{tabular}{l cc | c }
    \toprule
    Method  & Base Acc. & Novel Acc. & HM\\
    \midrule
    Independent V-L prompting (IVLP) & 84.21 & 71.79 & 77.51 \\
    PromptSRC with single prompt diversity & \textbf{84.32}	&75.52&	79.68 \\
    \rowcolor{tabhighlight}
    PromptSRC with ensembled prompt diversity & {84.26} & \textbf{76.10} & \textbf{79.97} \\
    \bottomrule
    \end{tabular}
    }\vspace{-0.5em}
    \caption{Analysis on alternate design choices for the textual diversity in PromptSRC. Incorporating textual diversity by ensembling multiple text templates achieves better generalization.
    }
    \label{tab_appendix:alternate_textual_diversity}
\end{table}

%% file: tables/supplementory/eva_clip_comparision.tex
\begin{table}[t!]
\centering
\tablestyle{-12pt}{1.1}
\addtolength{\tabcolsep}{+14pt}
\resizebox{0.8\columnwidth}{!}{%
\begin{tabular}{lc|c c c}
\toprule
{Dataset} &  &  {IVLP} &  {PromptSRC} &  {$\Delta$}  \\
\midrule

\multirow{3}{*}{\shortstack[l]{Average on\\  11 datasets}}       & Base Acc.      & {86.31} & \textbf{86.34}     &  \textcolor{MidnightBlue}{{+0.03}}\\
                               & Novel Acc.     & 74.96   & \textbf{78.68 }        &   \textcolor{MidnightBlue}{{+3.72}}\\
                               & HM          & 80.24 & \textbf{82.33 }       &   \textcolor{MidnightBlue}{{+2.09}}\\
\midrule
\multirow{3}{*}{ImageNet}      & Base Acc.    &  82.13 &  \textbf{82.40}           &  \textcolor{MidnightBlue}{{+0.27}}\\ 
                               & Novel Acc.      &  72.20 &\textbf{ 76.03 }     &   \textcolor{MidnightBlue}{{+3.83}}\\ 
                               & HM          &  76.85 & \textbf{79.09}     &  \textcolor{MidnightBlue}{{+2.24}}\\
\midrule
\multirow{3}{*}{Caltech101}    & Base Acc.   &  \textbf{99.33} & {98.97 }          &    \textcolor{Bittersweet}{{-0.36}}\\
                               & Novel Acc.       & 96.47 &\textbf{97.10 }& \textcolor{MidnightBlue}{{+0.63}}\\
                               & HM             &  97.88 &  \textbf{98.03} &  \textcolor{MidnightBlue}{{+0.15}}\\
\midrule
\multirow{3}{*}{OxfordPets}    & Base Acc.     & 95.17          & {95.63}  &  \textcolor{MidnightBlue}{{+0.46}}\\
                               & Novel Acc.       & {{98.43} }          & {98.43}  &   \textcolor{MidnightBlue}{{+0.00}}\\
                               & HM          & {96.77}          & \textbf{97.01}  &   \textcolor{MidnightBlue}{{+0.24}}\\
\midrule
\multirow{3}{*}{\shortstack[l]{Stanford \\ Cars}}  & Base Acc.       & \textbf{85.90}          & {85.07}  &  \textcolor{Bittersweet}{{-0.83}}\\
                               & Novel Acc.        & 83.97& \textbf{86.40 }          &  \textcolor{MidnightBlue}{{+2.43}}\\
                               & HM          & 84.92  & \textbf{ 85.73 }             &  \textcolor{MidnightBlue}{{+0.81}}\\
\midrule
\multirow{3}{*}{Flowers102}    & Base Acc.      & {99.47}  &{99.47  }        &  \textcolor{MidnightBlue}{{+0.00}}\\
                               & Novel Acc.       &77.43  &\textbf{79.57 }        & \textcolor{MidnightBlue}{{+2.14}}\\
                               & HM         & 87.08  & \textbf{88.41}      & \textcolor{MidnightBlue}{{+1.34}}\\
\midrule
\multirow{3}{*}{Food101}       & Base Acc.      & {90.60}       &\textbf{ 91.37 } &\textcolor{MidnightBlue}{{+0.77}}\\
                               & Novel Acc.      & 90.70  & \textbf{91.97  }       & \textcolor{MidnightBlue}{{+1.27}}\\
                               & HM           & 90.65  & \textbf{91.67 }      & \textcolor{MidnightBlue}{{+1.02}}\\
\midrule
\multirow{3}{*}{\shortstack[l]{FGVC \\ Aircraft}}  & Base Acc.   &\textbf{ 46.80}        & {46.40}  &  \textcolor{Bittersweet}{{-0.40}}\\
                               & Novel Acc.      & \textbf{28.90} &28.80          &  \textcolor{Bittersweet}{{-0.10}}\\
                               & HM           & \textbf{35.73}  & 35.54    &  \textcolor{Bittersweet}{-0.19}\\
\midrule
\multirow{3}{*}{SUN397}        & Base Acc.   &  83.30  &\textbf{84.50 }        &   \textcolor{MidnightBlue}{{+1.20}}\\
                               & Novel Acc.       &  76.93 &\textbf{80.80 }        &  \textcolor{MidnightBlue}{{+3.87}}\\
                               & HM          &  79.99 &\textbf{82.61}     &  \textcolor{MidnightBlue}{{+2.62}}\\
\midrule
\multirow{3}{*}{DTD}           & Base Acc.       & 84.60 &\textbf{86.27}             &  \textcolor{MidnightBlue}{{+1.67}}\\
                               & Novel Acc.       & 59.47 &\textbf{63.53  }      &  \textcolor{MidnightBlue}{{4.06}}\\
                               & HM            & 69.84 & \textbf{73.17}       & \textcolor{MidnightBlue}{{+3.33}}\\
\midrule
\multirow{3}{*}{EuroSAT}       & Base Acc.     & \textbf{96.13 } &93.43              &  \textcolor{Bittersweet}{{-2.70}}\\
                               & Novel Acc.        & 62.90& \textbf{82.30}           & \textcolor{MidnightBlue}{{+19.40}}\\
                               & HM             & 76.04& \textbf{87.51}   &  \textcolor{MidnightBlue}{{+11.47}}\\
\midrule
\multirow{3}{*}{UCF101}        & Base Acc.      & 86.00  &\textbf{86.23}           & \textcolor{MidnightBlue}{{+0.23}}\\
                               & Novel Acc.     & 77.20  &\textbf{80.57 }          &  \textcolor{MidnightBlue}{{+3.37}}\\
                               & HM           & 81.36 &\textbf{83.30}    & \textcolor{MidnightBlue}{{+1.94}}\\
\bottomrule
\end{tabular}
}\vspace{-0.5em}
    \caption{\small\textnormal{Compatibility of PromptSRC approach using a recent V-L model: EVA CLIP \cite{fang2023eva} in the Base-to-novel generalization setting}. PromptSRC shows overall favourable performance on EVA CLIP. Absolute gains over IVLP method are shown in \textcolor{MidnightBlue}{blue}.}
    \label{tab_appendix:eva_clip_comparison}
\end{table}

%% file: tables/supplementory/per_dataset_result_individual_components.tex
\begin{table*}[!t]
    \small \centering
 \setlength{\tabcolsep}{12pt}
    \scalebox{0.9}[0.9]{
\begin{tabular}{lc|cccc|c}
\toprule
\textbf{Dataset} & 
 & 
\textbf{IVLP} &
\textbf{+ $\mathcal{L}_\text{SCL}$} & 
\textbf{+ GPA} & 
\textbf{ + Textual diversity} &
\textbf{$\Delta$} \\ \midrule
\multirow{3}{*}{Average over 11 datasets}  & Base Acc.& 84.21       & 84.21   & 84.16  & 84.26 &   \textcolor{MidnightBlue}{{+0.04}}\\
                               & Novel Acc.     & 71.79  & 75.38   & 75.69  & 76.10  &   \textcolor{MidnightBlue}{{+4.31}}\\
                               & H.M            & 77.51  & 79.55   & 79.70 &79.97  &   \textcolor{MidnightBlue}{{+2.46}}\\
\midrule
\multirow{3}{*}{ImageNet}      & Base Acc.      & 77.00  & 77.53   &  77.47 &  77.60 &  \textcolor{MidnightBlue}{{+0.60}}\\ 
                               & Novel Acc.     & 66.50  & 69.77   &  70.03 & 70.73&    \textcolor{MidnightBlue}{{+4.23}}\\ 
                               & H.M            & 71.37  & 73.45   &  73.56 & 74.01&  \textcolor{MidnightBlue}{{+2.64}}\\
\midrule
\multirow{3}{*}{Caltech101}    & Base Acc.      & 98.30  & 98.03   & 97.97 & 98.10 &   \textcolor{Bittersweet}{{-0.20}}\\
                               & Novel Acc.     & 93.20  & 94.37   &  94.67 & 94.03 & \textcolor{MidnightBlue}{{+0.83}}\\
                               & H.M            & 95.68  & 96.17   &  96.29 &  96.02 &  \textcolor{MidnightBlue}{{+0.34}}\\
\midrule
\multirow{3}{*}{OxfordPets}    & Base Acc.      & 94.90  & 95.37   & 95.27 & {95.43}   &  \textcolor{MidnightBlue}{{+0.43}}\\
                               & Novel Acc.     & 97.20  & 97.03   & {97.10} & {97.30}  &   \textcolor{MidnightBlue}{{+0.10}}\\
                               & H.M            & 96.04  & 96.19   & {96.18} & {96.30}  &   \textcolor{MidnightBlue}{{+0.27}}\\
\midrule
\multirow{3}{*}{StanfordCars}  & Base Acc.     & 79.53 & 78.87    & 78.03 &78.27 &  \textcolor{Bittersweet}{{-1.26}}\\
                           & Novel Acc.         & 71.47 & 74.60    & 74.87&74.97   &  \textcolor{MidnightBlue}{{+3.50}}\\
                               & H.M            & 75.28 & 76.68    & 76.42  &  76.58 & \textcolor{MidnightBlue}{{+1.30}}\\
\midrule
\multirow{3}{*}{Flowers102}    & Base Acc.      & 97.97 & 97.97    & 98.00  &98.07&  \textcolor{MidnightBlue}{{+0.10}}\\
                               & Novel Acc.      & 72.10 & 76.90    & 77.10 &76.50 & \textcolor{MidnightBlue}{{+4.40}}\\
                               & H.M             & 83.07 & 86.17    & 86.30  & 85.95& \textcolor{MidnightBlue}{{+2.88}}\\
\midrule
\multirow{3}{*}{Food101}       & Base Acc.       & 89.37 & 90.37    & {90.57}  &90.67&\textcolor{MidnightBlue}{{+1.30}}\\
                               & Novel Acc.      & 90.30 & 91.23    & 91.47  & 91.53&  \textcolor{MidnightBlue}{{+1.23}}\\
                               & H.M            & 89.83 & 90.80    & 91.02  & 91.10& \textcolor{MidnightBlue}{{+1.27}}\\
\midrule
\multirow{3}{*}{FGVCAircraft}  & Base Acc.       & 42.60 & 42.33    & 42.30 & 42.73  &  \textcolor{MidnightBlue}{{+0.13}}\\
                               & Novel Acc.      & 25.23 & 35.60    & 36.83 &37.87   &  \textcolor{MidnightBlue}{{+12.6}}\\
                               & H.M             & 31.69 & 38.67    & 39.38  & 40.15  &  \textcolor{MidnightBlue}{{+8.46}}\\
\midrule
\multirow{3}{*}{SUN397}        & Base Acc.      & 81.60 & 82.53    &  82.57  &82.67 &   \textcolor{MidnightBlue}{{+1.07}}\\
                               & Novel Acc.      & 75.50 & 78.70    &  78.83 &78.47  &  \textcolor{MidnightBlue}{{+2.97}}\\
                               & H.M            & 78.43 & 80.57    &  80.66 &80.52  &  \textcolor{MidnightBlue}{{+2.08}}\\
\midrule
\multirow{3}{*}{DTD}           & Base Acc.       & 82.40 & 83.13    &82.97 &83.37   &  \textcolor{MidnightBlue}{{+0.97}}\\
                               & Novel Acc.      & 56.20 & 61.90    & 62.00 &62.97  &  \textcolor{MidnightBlue}{{+6.77}}\\
                               & H.M            & 66.82 & 70.96    & 70.97 &71.75  & \textcolor{MidnightBlue}{{+4.92}}\\
\midrule
\multirow{3}{*}{EuroSAT}       & Base Acc.       & 96.73 & 93.07    & 93.50  &92.90 &  \textcolor{Bittersweet}{{-3.83}}\\
                               & Novel Acc.      & 67.83 & 69.30    & 69.93& 73.90   & \textcolor{MidnightBlue}{{+6.07}}\\
                               & H.M            & 79.74 & 79.45    & 80.02& 82.32   &  \textcolor{MidnightBlue}{{+2.58}}\\
\midrule
\multirow{3}{*}{UCF101}        & Base Acc.      & 85.93  & 87.10   & 87.07  &87.10   & \textcolor{MidnightBlue}{{+1.17}}\\
                               & Novel Acc.      & 74.17  & 79.73   & 79.80  &78.80  &  \textcolor{MidnightBlue}{{+4.63}}\\
                               & H.M            & 79.62  & 83.25   & 83.28 &82.74  & \textcolor{MidnightBlue}{{+3.12}}\\
\bottomrule
\end{tabular}%
}
    \caption{\small Detailed performance comparison on individual datasets for showing effect of individual components in PromptSRC approach. Absolute gains of PromptSRC (IVLP + $\mathcal{L}_\text{SCL}$ + GPA + Textual diversity) over the IVLP are shown in \textcolor{MidnightBlue}{blue}. }
    \label{tab_appendix:base_to_new}
\end{table*}

%% file: tables/supplementory/few_shot_per_dataset.tex
\begin{table*}[t!]
    \small \centering
 \setlength{\tabcolsep}{15pt}
    \scalebox{0.9}[0.9]{
\begin{tabular}{ll|ccccc}
\toprule
\textbf{Dataset} & 
 \textbf{Method} & 
\textbf{1 shot} &
\textbf{2 shots} & 
\textbf{4 shots} & 
\textbf{ 8 shots} &
\textbf{16 shots} \\  \midrule

\multirow{4}{*}{ImageNet}      & Linear probe CLIP      &32.13	&44.88&	54.85	&62.23	&67.31\\ 
                               & CoOp                   &66.33	&67.07	&68.73	&70.63&	71.87\\ 
                               & CoCoOp                  & 69.43	&69.78	&70.39&	70.63&	70.83\\
                                                              & MaPLe                  & 62.67	&65.10	&67.70&	70.30&	72.33\\
                             \rowcolor{tabhighlight}   &  PromptSRC (Ours)              & 68.13	&69.77	&71.07	&72.33	&73.17\\
\midrule
\multirow{4}{*}{Caltech101}    & Linear probe CLIP      & 79.88	&89.01	&92.05	&93.41	&95.43\\
                               & CoOp                    & 92.60	&93.07	&94.40	&94.37	&95.57\\
                               & CoCoOp                  & 93.83	&94.82	&94.98	&95.04	&95.16\\
                                                              & MaPLe                  & 92.57	&93.97	&94.43&	95.20&	96.00\\
                               \rowcolor{tabhighlight} &  PromptSRC (Ours)                 &93.67	&94.53	&95.27	&95.67	&96.07\\
                               \midrule
\multirow{4}{*}{DTD}           & Linear probe CLIP       & 34.59	&40.76	&55.71	&63.46	&69.96\\
                               & CoOp                   &50.23&	53.60	&58.70	&64.77	&69.87\\
                               & CoCoOp                      & 48.54	&52.17	&55.04	&58.89	&63.04\\
                                                              & MaPLe                  & 52.13	&55.50	&61.00&	66.50&	71.33\\
                               \rowcolor{tabhighlight} &  PromptSRC  (Ours)                & 56.23&59.97	&65.53	&69.87	&72.73\\
                               \midrule
\multirow{4}{*}{EuroSAT}       & Linear probe CLIP       &49.23	&61.98	&77.09&	84.43	&87.21\\
                               & CoOp                   &54.93	&65.17	&70.80	&78.07	&84.93\\
                               & CoCoOp                  & 55.33	&46.74	&65.56&	68.21	&73.32\\
                                                              & MaPLe                  & 71.80	&78.30	&84.50&	87.73&	92.33\\
                              \rowcolor{tabhighlight}   & PromptSRC (Ours)             &73.13 &79.37	&86.30	&88.80	&92.43\\
\midrule
\multirow{4}{*}{StanfordCars}  & Linear probe CLIP     & 35.66	&50.28	&63.38	&73.67	&80.44\\
                           & CoOp                        &67.43	&70.50	&74.47	&79.30	&83.07\\
                                   & CoCoOp              & 67.22	&68.37	&69.39	&70.44&	71.57\\
                                                                  & MaPLe                  & 66.60	&71.60	&75.30&	79.47&	83.57\\
                              \rowcolor{tabhighlight}  &  PromptSRC (Ours)              &69.40	&73.40	&77.13	&80.97	&83.83\\
\midrule
\multirow{4}{*}{Flowers102}    & Linear probe CLIP      & 69.74	&85.07	&92.02	&96.10	&97.37\\
                               & CoOp                   & 77.53	&87.33	&92.17	&94.97	&97.07\\
                               & CoCoOp                    &72.08	&75.79	&78.40	&84.30	&87.84\\
                                                              & MaPLe                  & 83.30	&88.93	&92.67&	95.80&	97.00\\
                                \rowcolor{tabhighlight}  & PromptSRC (Ours)           & 85.93	&91.17	&93.87	&96.27	&97.60\\
                               \midrule
\multirow{4}{*}{FGVCAircraft}  & Linear probe CLIP       & 19.61	&26.41	&32.33	&39.35&	45.36\\
                               & CoOp                     & 21.37&	26.20	&30.83	&39.00	&43.40\\
                               & CoCoOp                   & 12.68	&15.06	&24.79	&26.61	&31.21\\
                                                              & MaPLe                  & 26.73	&30.90	& 34.87&	42.00&	48.40\\
                              \rowcolor{tabhighlight}   & PromptSRC (Ours)          & 27.67	&31.70	&37.47&	43.27	&50.83\\

                               \midrule
\multirow{4}{*}{SUN397}        & Linear probe CLIP          & 41.58	&53.70	&63.00	&69.08	&73.28\\
                               & CoOp                     & 66.77	&66.53	&69.97	&71.53	&74.67\\
                               & CoCoOp                   & 68.33	&69.03&	70.21	&70.84	&72.15\\
                                                              & MaPLe                  & 64.77	&67.10	&70.67&	73.23&	75.53\\
                             \rowcolor{tabhighlight}   &  PromptSRC (Ours)               &69.67	&71.60	&74.00	&75.73	&77.23\\
                               \midrule
\multirow{4}{*}{OxfordPets}    & Linear probe CLIP        & 44.06	&58.37	&71.17	&78.36	&85.34\\
                               & CoOp                      & 90.37	&89.80	&92.57	&91.27	&91.87\\
                               & CoCoOp                       & 91.27	&92.64	&92.81	&93.45	&93.34\\
                                                              & MaPLe                  & 89.10	&90.87	&91.90&	92.57& 92.83\\
                              \rowcolor{tabhighlight} &   PromptSRC (Ours)                  & 92.00	&92.50	&93.43	&93.50	&93.67\\
                               \midrule
\multirow{4}{*}{UCF101}        & Linear probe CLIP        & 53.66	&65.78	&73.28	&79.34&	82.11\\
                               & CoOp                     & 71.23	&73.43	&77.10	&80.20	&82.23\\
                               & CoCoOp                   & 70.30	&73.51	&74.82	&77.14&	78.14\\
                                                              & MaPLe                  & 71.83	&74.60	& 78.47& 81.37&	85.03\\
                             \rowcolor{tabhighlight}  &   PromptSRC (Ours)                  & 74.80	&78.50	&81.57	&84.30	&86.47\\
\midrule
\multirow{4}{*}{Food101}       & Linear probe CLIP       & 43.96	&61.51	&73.19	&79.79	&82.90\\
                               & CoOp                    & 84.33	&84.40	&84.47	&82.67	&84.20\\
                               & CoCoOp                     & 85.65	&86.22	&86.88	&86.97	&87.25\\
                                                              & MaPLe                  & 80.50 &81.47	&81.77&	83.60&	85.33\\
                              \rowcolor{tabhighlight} &   PromptSRC (Ours)               & 84.87&	85.70	&86.17	&86.90	&87.5\\

\midrule
\multirow{4}{*}{Average}       & Linear probe CLIP       & 45.83	&57.98	&68.01	&74.47&	78.79\\
                               & CoOp                    & 67.56	&70.65	&74.02	&76.98	&79.89\\
                               & CoCoOp                & 66.79&	67.65	&71.21&	72.96	&74.90\\
                                                              & MaPLe                  & 69.27	&72.58	&75.37&	78.89&	81.79\\
                               \rowcolor{tabhighlight} &  PromptSRC (Ours)              & 72.32	&75.29	&78.35	&80.69	&82.87\\

\bottomrule
\end{tabular}
}
    \caption{\small Per-dataset performance comparison of PromptSRC with various methods in few-shot setting.}
    \label{tab_appendix:few_shot_experiments}
\vspace{-5mm}
\end{table*}